\pdfoutput=1

\documentclass[11pt]{article}

\usepackage[preprint]{EMNLP2023}

\usepackage{times}
\usepackage{latexsym}
\usepackage{tabularx}
\usepackage{placeins}

\usepackage[T1]{fontenc}

\usepackage[utf8]{inputenc}

\usepackage{microtype}

\usepackage{inconsolata}
\usepackage{booktabs}
\usepackage{multirow}
\usepackage{array}
\usepackage{makecell}
\usepackage{graphicx}
\usepackage{amsfonts}
\usepackage[most]{tcolorbox}

\title{The Compressive Knowledge Graph Hypothesis: Which Graph Facts Matter for Scientific Hypothesis Generation?}

\author{
Shashwat Sourav$^{1,2,3,4}$,
Viktoriia Baibakova$^{2}$,
Sanjay Das$^{2}$,
Ran Elgedawy$^{2}$, \\
\textbf{Maria Mahbub}$^{2}$,
\textbf{Emily Herron}$^{2}$,
\textbf{Tirthankar Ghosal}$^{2}$ \\
\\
$^{1}$Washington University in St. Louis \\
$^{2}$Oak Ridge National Laboratory \\
$^{3}$Lawrence Berkeley National Laboratory \\
$^{4}$UniverseTBD \\
\texttt{s.shashwat@wustl.edu}
}

\begin{document}
\maketitle

\begin{abstract}
Knowledge graphs (KGs) can provide structured scientific context to language models, but it remains unclear which graph facts actually shape the generated hypotheses. We study KG-guided hypothesis generation for battery materials across Mistral-7B, Llama-3.1-70B, and Gemini 2.5 Flash. We perturb local KGs by varying density, ontology richness, topology, and control structure, and evaluate outputs with both provided-graph and fixed-reference metrics. Across models, KG utility is selective and model-dependent: graph context changes outputs, but no-KG outputs also recover substantial graph content from model priors. Compact top-$k$ subgraphs often approximate full-KG behavior, including when claimed-outcome triples are held out. At the same time, compression is not unique to one semantic ranking rule, random and topology-based subsets can also recover much of the signal. These results support a redundancy-aware Compressive KG hypothesis: useful KG signal is often recoverable from compact, scientifically structured subgraphs rather than requiring the full local graph. 
\end{abstract}

\section{Introduction}

Knowledge graphs are increasingly used as a way to give language models structured context \citep{2023arXiv230608302P, Ando2005}. In scientific settings, this is especially appealing \citep{2024arXiv241102382X,2025arXiv250504651K}. A graph can organize a problem into explicit concepts such as the material system, the failure mode, the proposed intervention, the mechanism, and the target property. In principle, this should help a model move from a vague answer to a more grounded scientific hypothesis \citep{2024arXiv240407738B}. In practice, however, it is still unclear how much of that graph structure the model actually uses \citep{2023arXiv230703172L, 2024arXiv241217031H}. A model may benefit from a few salient entities, from the relation structure itself, or from only a small subset of the graph while ignoring the rest. If we do not separate these possibilities, it is difficult to know what role external knowledge graphs are really playing in hypothesis generation.

Current work often treats knowledge graph prompting as a single intervention: provide the graph, then measure whether performance goes up or down \citep{2023arXiv230608302P,2023arXiv230809729W}. That view is too coarse for scientific discovery settings. A graph can vary in several ways at once. It can be dense or sparse, coarse or semantically rich, shallow or multi-hop \citep{2024arXiv240407103J,2024arXiv240520139M}. It can also be partly corrupted, shuffled, or compressed into a targeted subgraph \citep{2023arXiv230513269L}. When model behavior changes under these conditions, the main question is whether the graph helps, and which part of the graph helps, how that changes across models, and what kind of information remains useful as model capability increases \citep{2022arXiv220108860Z,2022arXiv221009338Y}.

In this work, we study these questions in a battery-science hypothesis-generation setting. We compare three models with different capability levels: Mistral-7B, Llama-3.1-70B, and Gemini 2.5 Flash \citep{2023arXiv231006825J,2024arXiv240721783G,2025arXiv250706261C}. For each scientific problem, we generate hypotheses under a family of KG conditions that vary density, ontology richness, topology, and control structure \citep{2024arXiv240407103J,2024arXiv240520139M,2024arXiv240710805M}. We then evaluate the generated outputs with metrics that separate entity use from relation use, and we add intervention-based analyses that test what information is necessary, sufficient, and causally important \citep{2024arXiv240208845C}. This lets us move beyond the usual question of whether KG prompting works, and toward a more precise account of how models use external structured knowledge.

Our main claim is what we call the Compressive Knowledge Graph Hypothesis. We propose that full-KG behavior is often recoverable from compact subgraphs rather than requiring the entire local graph. This is a redundancy-aware claim, as it does not require one ranking rule to uniquely identify the important triples. Instead, useful signal may be distributed across mechanism, intervention, failure-mode, and outcome-facing relations, and different compact subsets may preserve enough structure to recover similar hypothesis behavior \citep{a18010006,2025arXiv250216171L}.

Across analyses, we find that KG influence is real but model-dependent. Problem identity remains the dominant source of variation, while KG condition has a smaller but measurable effect. Targeted and top-$k$ subgraphs often recover much of the full-KG behavior; fixed-reference scoring shows that this is not merely an artifact of no-KG zeros; and an outcome-held-out control shows that compression is not only claimed-outcome leakage \citep{2024arXiv241020724L,2025arXiv250408893L}.



These findings matter for how knowledge graphs should be used in hypotheses generation. If the useful signal is concentrated in a compact subset, then supplying larger and denser graphs may not be the right design choice, especially for stronger models. A better approach may be to identify the small set of graph facts that actually steers generation. This has consequences not only for knowledge-graph prompting, but also for how we think about retrieval, structured context design, and evaluation in scientific language-model systems \citep{2023arXiv230108912G,2025arXiv250413079W}.

Our contributions are as follows:
\begin{itemize}
    \item We study knowledge-graph-guided hypothesis generation across three models and multiple graph manipulations, including density, ontology, topology, random, shuffled, targeted, and compressed graph conditions.
    \item We introduce an evaluation framework that separates provided-graph use from fixed-reference recovery, allowing no-KG, top-$k$, random, shuffled, and full-KG outputs to be scored against the same full graph.
    \item We provide evidence for a redundancy-aware \textbf{Compressive Knowledge Graph Hypothesis}: compact subgraphs often recover much of the full-KG behavior, including under an outcome-held-out control, but compression is not unique to a single semantic ranking rule.
\end{itemize}

\begin{figure*}[t]
    \centering
    \includegraphics[width=0.85\linewidth]{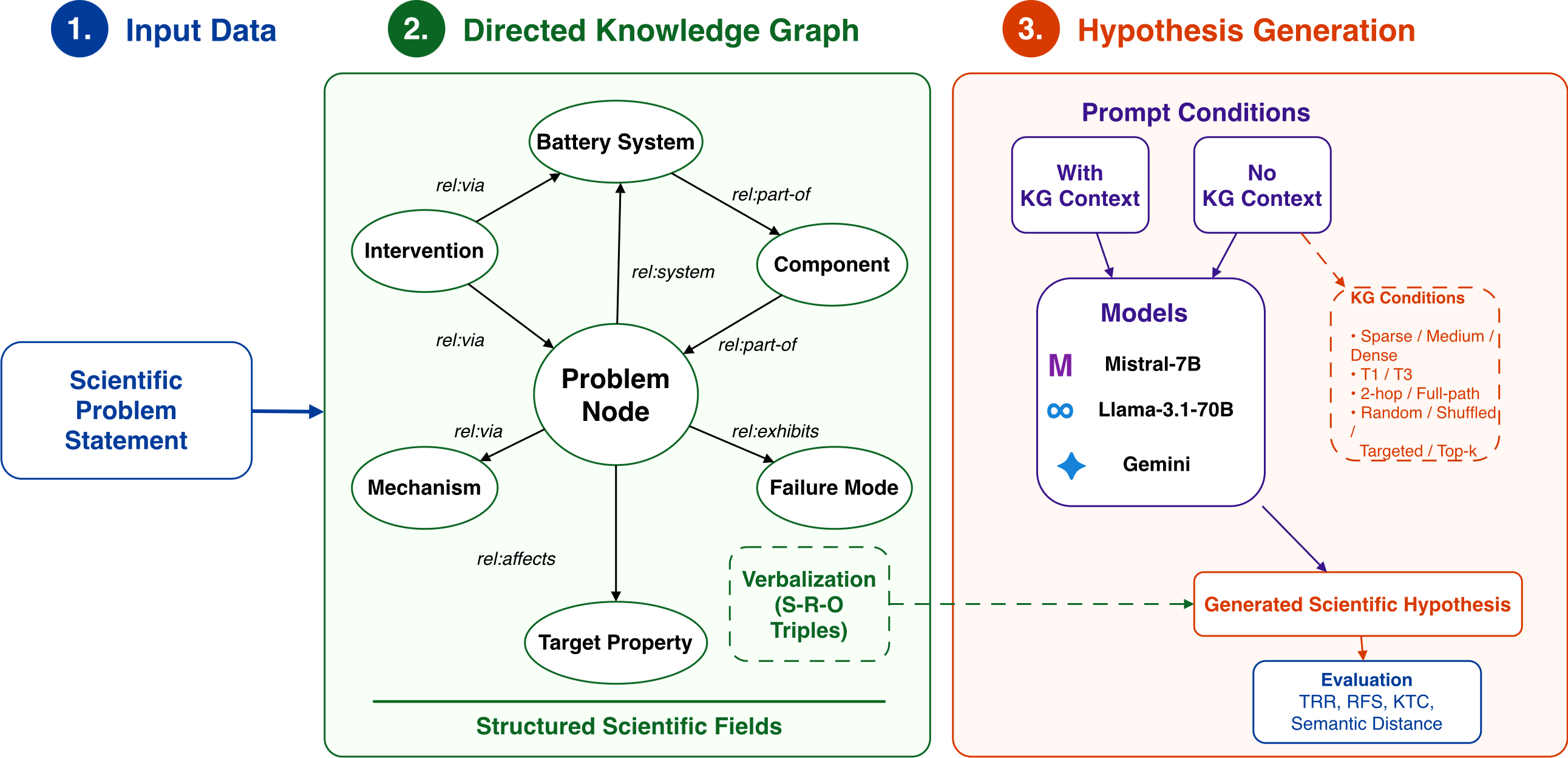}
\caption{\textbf{Overview of the KG-guided generation pipeline.}
Battery-science fields are converted into a directed KG, verbalized as triples, and provided to language models under different graph conditions. Outputs are evaluated for entity recall, relation fidelity, graph coverage, and semantic distance.}
    \label{fig:1}
\end{figure*}


\section{Related Work}
\label{sec:related_work}


Prior work has studied many ways of combining knowledge graphs with language models, including using KGs as external memory, structured prompts, reasoning scaffolds, or sources of factual grounding. Early knowledge-enhanced language-model work incorporated KG entities, triples, or verbalized KG facts into pretraining and representation learning \citep{2019arXiv190507129Z,2019arXiv190904164P,2020arXiv201012688A}. Surveys on knowledge-enhanced language models and LLM--KG integration argue that KGs can improve factuality, interpretability, and structured reasoning, while LLMs can help construct, complete, and verbalize KGs \citep{2022arXiv221105994H, 2023arXiv230608302P}. Recent GraphRAG work extends this idea by using graph-based indexing, graph-guided retrieval, and graph-enhanced generation to provide relational context for downstream tasks \citep{2024arXiv240808921P, 2025arXiv250100309H}. However, in our work, we are trying to understand which graph facts are actually used during scientific hypothesis generation.



Previous works have also developed methods that use KGs to guide hypothesis generation. KG-CoI integrates external structured knowledge into a chain-of-ideas process and uses KG support to reduce hallucinations in hypotheses generation \citep{2024arXiv241102382X}. Related systems use scientific KGs for link prediction, literature-based discovery, or candidate hypothesis ranking \citep{10.1145/2623330.2623667,pu2023graph,10.1016/j.knosys.2025.113280,2025arXiv250612385K}. These approaches generally treat the KG as useful context or evidence. By contrast, we test the compressive view of KG utility: whether a small subset of high-value triples is sufficient to recover much of the full-KG behavior, whether removing that subset disrupts generation, and whether triple importance is governed by semantic role rather than graph topology alone.

\section{Task and Experimental Setup}
\label{sec:setup}

We study KG-guided hypothesis generation in battery science \citep{CHEN2026101091}. Each example contains a scientific problem statement and structured fields such as material system, component, failure mode, intervention, mechanism, target property, and claimed outcome. In total we used 100 problems.\footnote{The 100-problem evaluation dataset is available at \url{https://huggingface.co/datasets/matter2mech/battery-science-problems}.} From these fields, we construct a directed KG whose typed triples connect the problem to relevant scientific concepts. The model is asked to generate a solution hypothesis either without graph context or with a verbalized set of subject-relation-object triples, a common strategy for injecting KG facts into language-model prompts \citep{2019arXiv190907606L,2020arXiv201012688A,2024arXiv240407738B}.

All KG variants are derived from the same local graph $G_p=(V_p,E_p)$, with 15-18 typed triples per problem. Density variants vary subset size; ontology variants vary relation granularity; topology variants range from 2-hop context to full problem-to-outcome paths. Random, shuffled, and targeted/top-$k$ controls respectively test irrelevant triples, broken relations, and compact relevance-ranked subsets. Full definitions and size/context statistics are given in Appendices~\ref{app:kg_variant_definitions} and~\ref{app:kg_size}.

Battery materials are a useful testbed because the problems are inherently relational and mechanism-driven. A good hypothesis must connect what material or component is involved, why it fails, how an intervention changes the mechanism, and which property or outcome should improve. In material science, the hypothesis must preserve links among degradation, ion transport, interfacial stability, capacity retention, and materials design. These recurring structures are central to battery research and can be represented as KG triples, making the domain well suited for testing whether models use graph relations rather than only surface entities \citep{2025arXiv250113299K}.

\begin{figure*}[t]
    \centering
    \includegraphics[width=\linewidth]{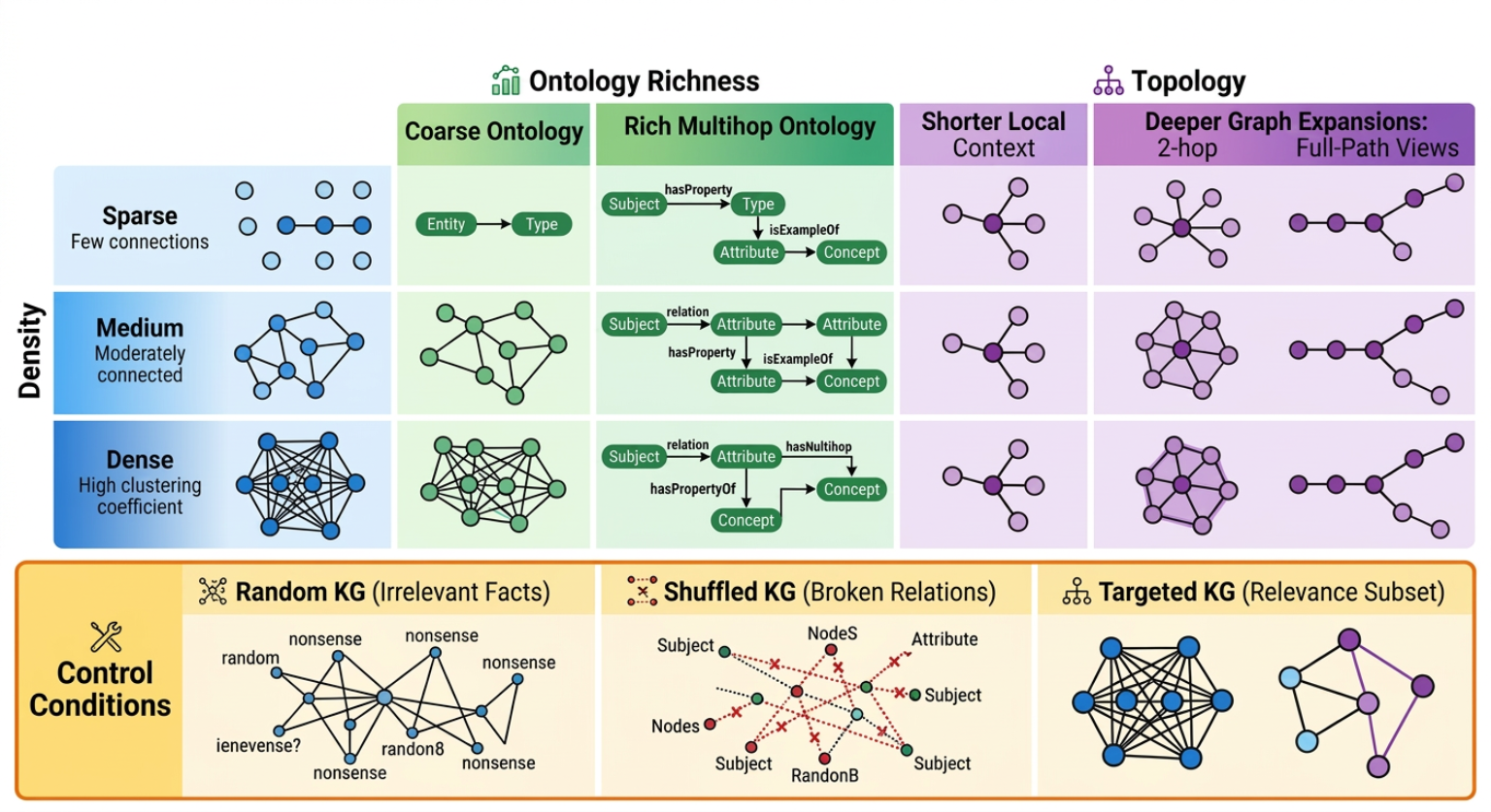}
    \caption{
    \textbf{KG perturbation design.}
    We vary the external knowledge graph along three axes: density, ontology richness, and topology. Density controls how many graph facts are supplied; ontology richness controls whether relations are coarse or semantically detailed; topology controls whether the model receives local 2-hop context or longer full-path structure.
    }
    \label{fig:kg_design}
\end{figure*}

We compare Mistral-7B, Llama-3.1-70B, and Gemini 2.5 Flash in the main cross-family study, and run additional intra-family checks on Mistral-7B/12B/22B and Llama-3.1-8B/70B. For each problem, we keep the prompt fixed and vary only the graph condition. The KG manipulations span three axes (\ref{fig:kg_design}): density (sparse, medium, dense), ontology richness (coarse T1 versus richer T3 multihop relations), and topology (2-hop versus full-path context). We also include control conditions: no KG, random KG, shuffled KG, targeted KG, and top-$k$ compressed subgraphs. Detailed definitions of the random, shuffled, entity-only, and relation-skeleton controls are given in Appendix~\ref{app:kg_controls}.

We test three claims. First, real KG structure should affect outputs more than irrelevant or corrupted graph context. Second, if KG utility is compressed, then a small top-$k$ subset should recover much of the full-KG behavior, while removing that subset should degrade the output. Third, graph influence should not be fully explained by simple topology alone: relation type, task relevance, and redundancy among local triples may all shape which compact subsets recover full-KG behavior.

\section{Evaluation Metrics}
\label{sec:metrics}

We design the evaluation to measure whether a knowledge graph changes the generated hypothesis, how it changes it. This distinction is important because a model may copy graph entities without using their relations, preserve relation language without recalling the correct objects, or generate a hypothesis that is semantically close to the full-KG output while covering only a small fraction of the graph. We therefore use a set of output-based metrics that separate entity recall, relation use, graph coverage, and semantic sensitivity. We distinguish two scoring views. Provided-graph metrics score an output against the graph supplied in that condition and measure whether the model used the given graph. These metrics are undefined for no-KG settings and can have different denominators for top-$k$ and full-KG conditions. We therefore also report fixed-reference metrics, which score every output against the same full KG for that problem. Fixed-reference scoring makes no-KG, random, shuffled, top-$k$, and full-KG outputs directly comparable and avoids artificial zero scores for no-KG baselines. 

\paragraph{Human expert evaluation.}
Automatic graph-use metrics can show whether an output reflects KG content, but they do not fully measure whether the hypothesis is scientifically useful. We therefore add a blinded domain-rater assessment. A materials-science postdoctoral researcher rated five representative examples comparing No KG, Top-8 KG, and Full KG outputs. The outputs were anonymized and condition labels were hidden during rating. The rater scored each hypothesis on a 1-5 scale for problem alignment, mechanistic specificity, intervention specificity, scientific plausibility, and evidence faithfulness, and also gave pairwise preferences against the No-KG output. The full rating instructions are provided in Appendix~\ref{app:human_eval_instructions}.



\paragraph{Triple Recall Rate.}
Triple Recall Rate (TRR) measures how much of the provided KG content appears in the final hypothesis at the object-entity level. For a graph condition with triples $\mathcal{G}=\{(s_i,r_i,o_i)\}_{i=1}^{n}$ and generated hypothesis $y$, we define
\[
\mathrm{TRR}(y,\mathcal{G}) =
\frac{1}{|\mathcal{O}_{\mathcal{G}}|}
\sum_{o \in \mathcal{O}_{\mathcal{G}}}
\mathbf{1}[o \in y],
\]
where $\mathcal{O}_{\mathcal{G}}$ is the set of object entities in the supplied graph. TRR captures whether the model recalls the entities made available by the graph. However, TRR alone does not show whether the model used the graph structure correctly, since an output can mention the right entities while ignoring or distorting their relations.

\paragraph{Relation Fidelity Score.}
Relation Fidelity Score (RFS) measures whether the generated hypothesis preserves the semantic role of the KG relations. Each graph relation is mapped to a scientific entity such as failure mode, mechanism, intervention, material component, property, or outcome. RFS then measures whether the language of the generated hypothesis expresses the same relation type. This allows us to distinguish shallow entity copying from relation-aware graph use. For example, mentioning an electrolyte additive contributes to TRR, but it contributes to RFS only if the output uses it in a role consistent with the supplied graph relation, such as an intervention that stabilizes an interface or improves ionic transport.

\paragraph{KG Triple Coverage.}
KG Triple Coverage (KTC) measures broader coverage of the graph content. While TRR focuses on object entities, KTC measures the fraction of supplied triples whose object-side content is represented in the generated hypothesis. This metric is useful for comparing full-KG, no-KG, random-KG, and compressed-KG conditions because it directly measures how much of the graph context is reflected in the output. A high KTC score indicates that the model uses a larger portion of the graph, whereas a low KTC score indicates that the model either ignores the graph or uses only a small subset of it.

\paragraph{Fixed-reference graph recovery.}

Provided-graph metrics answer whether a model used the graph it received. To compare conditions with different graph sizes, we also compute fixed-reference variants. Let $\mathcal{G}_{p}^{\mathrm{full}}$ be the full KG for problem $p$. For any condition $c$, including no KG, random KG, shuffled KG, and top-$k$, we compute
\[
\mathrm{TRR}_{\mathrm{ref}}(y_{p,c}) =
\mathrm{TRR}(y_{p,c}, \mathcal{G}_{p}^{\mathrm{full}}).
\]
Analogous fixed-reference versions are computed for relation fidelity and graph coverage. These metrics ask how much of the full problem graph is recovered in the output, regardless of which graph was provided to the model.

\paragraph{Semantic distance to full-KG behavior.}

For sufficiency and comprehensiveness experiments, we also measure semantic distance between an ablated output and the corresponding full-KG output. Let $y_{\mathrm{full}}$ be the hypothesis generated with the full KG and $y_c$ be the hypothesis generated under condition $c$, such as top-$k$ triples or full KG with top-$k$ triples removed. We compute
\[
d_{\mathrm{sem}}(y_c,y_{\mathrm{full}}) =
1 - \cos\left(e(y_c), e(y_{\mathrm{full}})\right),
\]
where $e(\cdot)$ is a sentence embedding function. Lower semantic distance means that the ablated condition better recovers the behavior induced by the full graph. This is our main metric for testing whether a small subset of triples is sufficient to approximate full-KG behavior. Additional metric implementation details are provided in Appendix~\ref{app:implementation}.


\paragraph{Intra-problem versus inter-problem variation.}

To measure how strongly graph condition affects generation relative to the problem itself, we compare two sources of semantic variation. Intra-problem variation measures how much outputs change for the same scientific problem when the KG condition changes. Inter-problem variation measures how much outputs differ across different scientific problems. We summarize this using the variance ratio
\[
\rho =
\frac{\mathbb{E}_{p}\left[d(y_{p,c}, y_{p,c'})\right]}
{\mathbb{E}_{p \neq p'}\left[d(y_{p,c}, y_{p',c'})\right]},
\]
where $p$ indexes problems and $c,c'$ index KG conditions. Smaller values of $\rho$ indicate that problem identity dominates over graph condition, while larger values indicate stronger sensitivity to graph context.

\paragraph{Statistical testing.}
We use paired permutation tests for the main condition contrasts because each problem is evaluated under multiple KG conditions. We also report bootstrap confidence intervals for key deltas, including $\Delta$TRR(real-random), $\Delta$RFS(real-shuffled), and fixed-reference recovery differences. Because we test multiple model-metric-condition contrasts, we report both uncorrected and Holm/BH-corrected p-values in Appendix~\ref{app:stats}. We show numerical zeros as $p<0.0001$ and near-one values as $p>0.999$. The main text emphasizes effect sizes, confidence intervals, and consistent directional patterns rather than isolated significance thresholds.

Implementation details for RFS, KTC, and the deterministic top-$k$ triple-ranking rule are provided in Appendix~\ref{app:implementation}, with additional details in Appendices~\ref{app:rfs_ktc} and~\ref{app:topk_ranking}. We compute semantic distance with a Sentence-Transformers encoder and verify (Appendix~\ref{app:encoder_sensitivity}) that the sufficiency trend is robust to replacing MiniLM-L6 with MPNet-base. The two encoders give highly correlated distances (Spearman $\rho=0.965$) and preserve the same monotonic decrease from $k=1$ to $k=8$.


\begin{table*}[t]
\centering
\small
\setlength{\tabcolsep}{6pt}
\caption{Model-level summary of KG utility. $\Delta$TRR(real$-$random) measures sensitivity to replacing the real KG with a random KG; $\Delta$RFS(real$-$shuffled) measures sensitivity to shuffling KG structure; $\Delta$KTC(real$-$noKG) measures the gain from using the real KG over no KG context. The variance ratio is intra-problem semantic variation across KG conditions divided by inter-problem variation across scientific problems; lower values indicate that problem identity dominates more strongly over KG condition.}
\label{tab:summary_main}
\begin{tabular}{lccccccc}
\toprule
Model & $\Delta$TRR & $\Delta$RFS & $\Delta$KTC & Variance ratio & Best density & Best ontology & Best topology \\
\midrule
Gemini         & 0.2900 & 0.1867  & 0.7569 & 0.3125 & sparse & T3\_multihop & full\_path \\
Llama-3.1-70B  & 0.0380 & 0.0100  & 0.0621 & 0.3838 & sparse & T1\_coarse   & 2hop \\
Mistral-7B     & 0.0080 & -0.0800 & 0.0054 & 0.5148 & dense  & T3\_multihop & 2hop \\
\bottomrule
\end{tabular}
\end{table*}

\section{Results}
\label{sec:results}

Our results support the redundancy-aware Compressive Knowledge Graph Hypothesis. Across all our experiments, we observe three main patterns. First, KG utility is model-dependent. Second, compact subgraphs often approximate full-KG behavior, including when claimed-outcome triples are held out. Third, graph influence is not explained by a single selector: relation role matters, but random and topology-based subsets can also recover much of the signal when enough triples are retained.

\subsection{Cross-family KG utility is selective}
\label{sec:cross_family}

Table~\ref{tab:summary_main} summarizes the main cross-family results. Gemini shows the strongest response to real KG structure, with $\Delta$TRR(real--random) = 0.290 and $\Delta$KTC(real--noKG) = 0.757. Llama-3.1-70B shows a smaller but positive effect, especially in graph coverage. Mistral-7B shows weak and brittle graph use, with almost no gain over random KG context and negative $\Delta$RFS under the real-versus-shuffled comparison.

The preferred graph structure also differs by model. Gemini 2.5 Flash benefits most from sparse, semantically rich, full-path graph context. Llama-3.1-70B benefits most from sparse, coarse, shorter-range context. Mistral-7B prefers denser scaffolding. This supports the selective-utility view: stronger models do not necessarily need more graph context, but they can benefit from a smaller set of high-signal facts. Full permutation tests and bootstrap confidence intervals are reported in Appendix~\ref{app:stats}.

We observe this in both the control comparisons and the structural ablations. In Table~\ref{tab:perm_controls}, Gemini 2.5 Flash shows large and statistically stable gains when the real KG is compared against a random KG, especially for TRR and KTC. It also shows a strong gain in RFS when the real KG is compared against a shuffled KG, indicating that relation structure is being used in a meaningful way. Llama-3.1-70B shows a smaller but still positive pattern, with the clearest gain appearing in KTC under the real-versus-random comparison. By contrast, Mistral-7B shows little benefit relative to the random KG and is strongly degraded by shuffled structure, suggesting that its graph use is weak and brittle rather than robust. A mixed-effects variance decomposition shows that model identity explains more variance than KG condition across TRR, RFS, and KTC; we report the full table in Appendix~\ref{app:stats}.

These results show that external KG utility does not vanish across models, but it becomes more selective. Stronger models do not benefit uniformly from larger or richer graphs. Instead, they appear to benefit most from smaller, higher-signal graph contexts.

\paragraph{Materials-science prompting guidance.}
The best graph condition differs by model, but the top examples share a common materials-science structure: the graph preserves a chain from failure mode to intervention, mechanism, target property, and outcome. Gemini 2.5 Flash performs best with sparse, semantically rich full-path graphs, suggesting that it can use compact mechanism-to-outcome chains. Llama-3.1-70B performs best with sparse coarse 2-hop graphs, suggesting that concise local grounding is sufficient. Mistral-7B benefits more from dense rich 2-hop graphs, suggesting that smaller open models may need more explicit local scaffolding. Representative top graph examples for each model are shown in Appendix~\ref{app:top_graph_examples}.

\subsection{Compressed subgraphs recover full-KG behavior}
\label{sec:compression}

We next test whether the full graph is necessary. The necessity analysis compares no KG, entity-only context, relation skeletons, targeted KG, and full KG; full results are reported in Appendix~\ref{app:necessity_controls}. Entity-only context does not recover the full-KG effect, showing that graph utility is not just lexical exposure to scientific terms. Relation skeletons preserve some relation-level signal but lose entity-specific grounding. Targeted KG recovers much of the full-KG behavior, indicating that useful graph signal can be preserved by compact subsets rather than requiring the entire graph. Appendix~\ref{app:ranking_baselines} further shows that this compression effect is not unique to our semantic ranking heuristic: random and topology-based top-$k$ selectors also approach full-KG behavior as $k$ increases, although no single selector dominates across all models and $k$ values.

\begin{figure*}[t]
    \centering
    \includegraphics[width=0.9\linewidth]{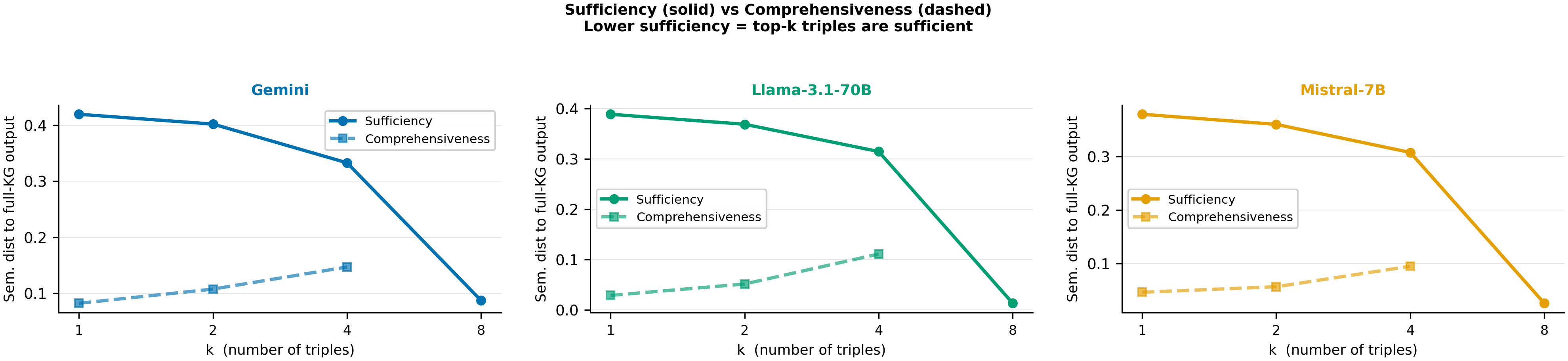}
    \caption{
    \textbf{Sufficiency and comprehensiveness support KG compression.}
    Keeping only the top-$k$ ranked triples increasingly recovers the full-KG output, while removing the same triples causes systematic degradation. Compact subsets are often sufficient to approximate full-KG behavior, while removing high-ranked subsets causes systematic degradation.
    }
    \label{fig:suf_comp}
\end{figure*}

The top-$k$ sufficiency and comprehensiveness analyses test this directly. In the sufficiency setting, we keep only the top-$k$ ranked triples and measure distance to the full-KG output. In the comprehensiveness setting, we remove the same top-$k$ triples and measure degradation. Figure~\ref{fig:suf_comp} shows the key pattern: as $k$ increases, top-$k$ triples become increasingly sufficient to recover full-KG behavior, while removing them causes larger disruption. This is the central empirical signature of the Compressive Knowledge Graph Hypothesis.

Because our top-$k$ ranking uses lexical overlap with the problem statement, we also test whether compression is merely an artifact of this heuristic. We compare semantic top-$k$ subsets against matched random-$k$ and topology-based subsets selected by degree, betweenness, and PageRank. The compression trend remains across ranking methods, so our main claim is not that the lexical ranking is optimal, but that full-KG behavior can be approximated by compact subsets of triples. Full results are reported in Appendix~\ref{app:ranking_baselines}.

The expert scores in Table~\ref{tab:expert_case_study} support the compression pattern. No-KG outputs are often plausible, but they are less mechanistically specific and less faithful to the supplied evidence. Top-8 KG nearly matches Full KG on mechanistic specificity and evidence faithfulness, while keeping scientific plausibility high. Because the rater saw anonymized outputs without condition labels, this provides a small sanity check that compact graph context recovers useful scientific grounding rather than merely increasing surface overlap. Full rating instructions are given in Appendix~\ref{app:human_eval_instructions}. On average, each per-problem KG contains 16.1 triples (median 16, range 15-18), so the top-8 subset corresponds to only 49.7\% of the full local graph and top-4 to 24.8\%. Full KG-size and context-length statistics are reported in Appendix~\ref{app:kg_size}. This shows that the observed recovery is not due to using nearly the whole graph, but to a compact subset of graph facts.


\paragraph{Outcome-held-out control.}

Because the local KG includes claimed-outcome triples, we test whether compression is merely an artifact of exposing part of the target hypothesis. We remove all outcome-facing triples and rerun full-KG and top-$k$ conditions for Gemini and Llama-3.1-70B. Table~\ref{tab:outcome_holdout} shows that removing outcome triples reduces relation fidelity, as expected, but does not eliminate graph signal. For Gemini, full-KG RFS drops from 0.580 to 0.439, while top-8 without outcome triples still reaches 0.458. For Llama-3.1-70B, full-KG RFS drops from 0.677 to 0.511, while top-8 without outcome triples reaches 0.550. Mechanism/intervention coverage is also preserved under top-8 no-outcome conditions. Thus, outcome-facing triples are high-leverage, but the compression effect is not only outcome leakage.

\begin{table}[t]
\centering
\small
\setlength{\tabcolsep}{4pt}
\caption{
\textbf{Outcome-held-out control.}
Removing claimed-outcome triples reduces relation fidelity but does not eliminate graph signal. Compact top-8 subgraphs without outcome triples preserve substantial relation and mechanism/intervention signal.
}
\label{tab:outcome_holdout}
\begin{tabular}{llcc}
\toprule
Model & Condition & RFS & Mech./Int. \\
\midrule
\multirow{3}{*}{Gemini} 
& Full KG           & 0.580 & 0.075 \\
& Full KG -- out.   & 0.439 & 0.075 \\
& Top-8 -- out.     & 0.458 & 0.073 \\
\midrule
\multirow{3}{*}{Llama-70B}
& Full KG           & 0.677 & 0.107 \\
& Full KG -- out.   & 0.511 & 0.093 \\
& Top-8 -- out.     & 0.550 & 0.110 \\
\bottomrule
\end{tabular}
\end{table}



\begin{table}[t]
\centering
\small
\setlength{\tabcolsep}{3.5pt}
\caption{
\textbf{Human Expert Evaluation}
A materials-science postdoc blindly scored five representative examples on a 1--5 scale. Top-8 KG closely tracks Full KG on mechanism and evidence grounding, while both improve over No KG.
}
\label{tab:expert_case_study}
\begin{tabular}{lccc}
\toprule
Criterion & No KG & Top-8 KG & Full KG \\
\midrule
Problem alignment & 3.2 & 4.0 & 4.2 \\
Mechanistic specificity & 2.4 & 3.8 & 4.0 \\
Intervention specificity & 2.6 & 3.6 & 3.8 \\
Scientific plausibility & 3.8 & 3.8 & 4.0 \\
Evidence faithfulness & 2.0 & 3.8 & 4.2 \\
\midrule
Pairwise vs No KG & -- & 4/5 & 5/5 \\
Top-8 close to Full & -- & 4/5 & -- \\
\bottomrule
\end{tabular}
\end{table}

\subsection{Topology alone does not explain graph influence}
\label{sec:semantic_role}

Compression alone does not identify which triples matter. We therefore use contradiction, knockout, and ranking-baseline experiments to test whether graph influence is driven by relevance, relation type, topology, or redundancy. In contradiction experiments, high-relevance incorrect triples cause larger semantic shifts than low-relevance contradictions and reduce TRR more strongly. This shows that models are not treating the KG as passive background context: corrupting a central fact can change the generated hypothesis.

Relation-type analysis shows that outcome-facing and task-relevant triples are often influential, but topology-based selectors and random subsets can also recover full-KG behavior when enough triples are retained. This suggests that triple importance is not reducible to any single ranking rule; rather, the full graph contains redundant routes to similar hypothesis behavior, and compact subsets can preserve enough of that signal. Additional knockout statistics for bridge, peripheral, random, and relation-type removals are shown in Appendix~\ref{app:topology_semantics}.

\begin{figure*}[htbp]
    \centering
    \includegraphics[width=0.99\linewidth]{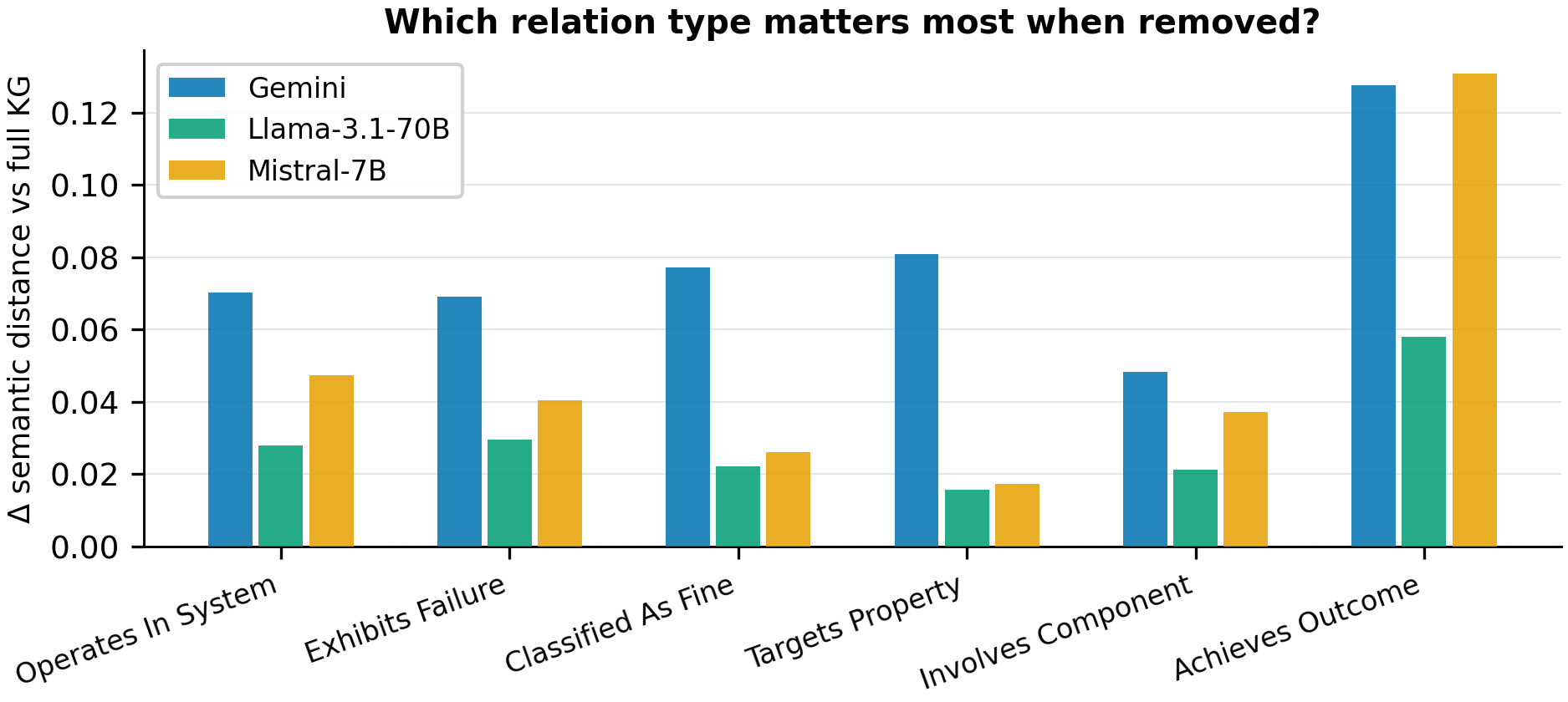}
    \caption{
    \textbf{Triple importance depends on semantic role.}
    Removing triples grouped by relation type shows that the most disruptive graph facts are not determined by topology alone. Outcome-facing and task-relevant relations often produce larger changes than purely structural bridge status, supporting the view that the compressed useful subset is semantically organized.
    }
    \label{fig:knockout_relation}
\end{figure*}






\subsection{Intra-family scaling shows compression without a simple size law}
\label{sec:within_family}

The cross-family comparison could reflect differences in training data, alignment, or model family rather than scale alone. We therefore perform intra-family experiments for Mistral and Llama. In the Mistral family, semantic distance to the full-KG output decreases as $k$ increases for 7B, 12B, and 22B models, confirming that compressed subsets recover full-KG behavior. However, the trend is not monotonic in parameter count. The 22B model is not consistently closer to full-KG behavior than the 12B or 7B models.

\begin{figure}[htbp]
    \centering
    \includegraphics[width=\linewidth]{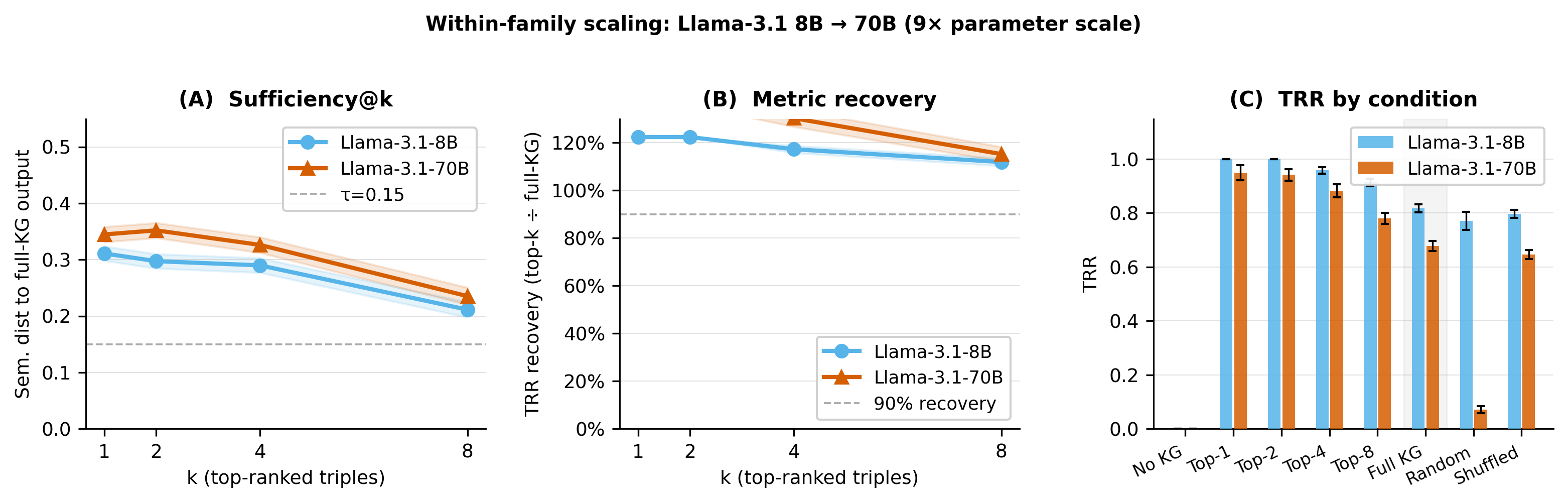}
    \caption{Within-family scaling for Llama-3.1 8B$\rightarrow$70B.
    Panel (A) shows semantic distance between top-$k$ outputs and the full-KG output; lower values mean that the compressed subset better recovers full-KG behavior. Panel (B) shows TRR recovery relative to the full-KG condition. Panel (C) compares TRR across KG conditions. Both models show compression, but the 70B model is less driven by broad graph recall under full, random, and shuffled graph contexts.
    }
    \label{fig:llama_scaling}
\end{figure}

Figure~\ref{fig:llama_scaling} shows the same compression pattern within Llama-3.1 8B and 70B. Both models move closer to full-KG behavior as more top-ranked triples are added. The 70B model is less driven by broad graph recall under full, random, and shuffled settings. Thus, scaling changes the form of KG dependence, but does not produce a simple rule that larger models always need fewer triples. The full Mistral-family sufficiency and TRR tables are reported in Appendix~\ref{app:within_family}. We also studied whether these effects are sensitive to generation randomness. On a repeated-sampling subset, between-condition TRR differences are much larger than within-condition sampling variation. About $18.1\times$ larger for Llama-3.1-70B and $14.7\times$ larger for Mistral-7B. This shows that the main effects are not driven by sampling noise. Detailed repeated-sampling statistics are reported in Appendix~\ref{app:sampling_variability}. Encoder-sensitivity results are reported in Appendix~\ref{app:encoder_sensitivity}. Additional knockout statistics for bridge, peripheral, and random removals are shown in Appendix~\ref{app:topology_semantics}.


\section{Discussion}
\label{sec:discussion}


We find that external graphs influence hypothesis generation, but their useful signal is not proportional to graph size. Compact subgraphs often approximate full-KG behavior, and removing high-ranked subsets can disrupt generation. However, the revised fixed-reference and ranking-baseline analyses show that this compression should be interpreted as redundancy-aware rather than selector-unique. No-KG outputs already recover some full-KG entities from model priors, and random or topology-based compact subsets can also recover much of the full-KG signal when enough triples are retained.

The outcome-held-out control further shows that compression is not merely claimed-outcome leakage. Removing outcome-facing triples reduces relation fidelity, but top-8 no-outcome subgraphs still preserve substantial relation and mechanism/intervention signal. Thus, outcome triples are high-leverage, but useful KG signal also flows through mechanism, intervention, and failure-mode relations. These findings suggest that graph compression should be guided by scientific role, but not only by role. Failure modes, interventions, mechanisms, and outcomes are useful signals, yet fixed-reference baselines show that redundancy and model priors also matter. The practical implication is that KG-guided generation should diagnose which compact pieces of context change behavior, rather than assuming that larger graph neighborhoods are always better.

\section{Conclusion}
We studied how language models use external knowledge graphs for hypotheses generation. Across Mistral-7B, Llama-3.1-70B, and Gemini 2.5 Flash, KG utility is model-dependent. Compact subgraphs often approximate full-KG behavior, and outcome-held-out results show that this compression is not just claimed-outcome leakage. At the same time, fixed-reference and ranking-baseline analyses show that compression is not unique to one semantic selector; local scientific KGs contain redundant routes to similar hypothesis behavior. Our results suggest that KG-guided hypothesis generation should focus less on expanding graph size alone and more on diagnosing which compact pieces of structured context actually change model behavior.


\section{Limitations}
\label{sec:limitations}

This study is limited to hypothesis generation in materials science, so the results may not be directly transferred to other scientific domains. Our graph-use metrics are diagnostic rather than full measures of scientific value. Fixed-reference scoring reduces denominator artifacts and avoids artificial no-KG zeros, but it still measures recovery of structured graph content rather than expert judgments of novelty, feasibility, or experimental usefulness. We therefore interpret the results as evidence about how models use structured context, not as a complete evaluation of hypothesis quality.

\section{Acknowledgment}

S.S, R.E, T.G acknowledge support from the U.S. Department of Energy, Office of Science, Office of Advanced Scientific Computing Research and Office of Basic Energy Sciences, Scientific Discovery through Advanced Computing (SciDAC) program under the FORUM-AI project. E.H., S.D. acknowledges support from the Oak Ridge Leadership Computing Facility (OLCF), which is a DOE Office of Science User Facility at the Oak Ridge National Laboratory supported by the U.S. Department of Energy under Contract No. DE-AC05-00OR22725. V.B. acknowledges support from US DOE Roll-to-Roll Consortium.

\FloatBarrier
\bibliography{custom}

\FloatBarrier
\appendix


\section{Qualitative Case Study}
\label{app:case_study}

Table~\ref{tab:case_study} shows one representative example. The no-KG output gives a plausible but generic additive-based hypothesis. The full-KG output is more specific: it identifies a low-molecular-weight fluorinated ether additive, connects it to dense electrical double layer ordering, and links that mechanism to desolvation impedance. This illustrates why entity recall alone is insufficient: the important change is the intervention--mechanism--failure connection.

\subsection{Human expert Assessment Instructions}
\label{app:human_eval_instructions}

We asked one materials-science postdoc to rate five representative examples. Each example contained the problem context and anonymized hypotheses from the No KG, Top-8 KG, and Full KG conditions. The condition labels were hidden during rating. The rater was instructed not to infer which system produced which output and to judge the hypotheses only by their scientific content. The rater was recruited through professional research contact and was not recruited through a crowdsourcing platform; no paid crowdwork was used.

The rater received the following instructions:

\begin{quote}
For each example, please read the problem/context and the anonymized hypotheses. Please rate each hypothesis on a 1--5 scale, where 1 means poor or absent, 3 means acceptable, and 5 means strong.

\textbf{Problem alignment:} Does the hypothesis address the stated battery/materials problem?

\textbf{Mechanistic specificity:} Does the hypothesis give a concrete mechanism rather than generic improvement language?

\textbf{Intervention specificity:} Does the hypothesis name a concrete material, process, coating, additive, architecture, or design intervention?

\textbf{Scientific plausibility:} Is the hypothesis chemically or materials-science plausible?

\textbf{Evidence faithfulness:} Is the hypothesis faithful to the supplied structured evidence and graph facts?

After scoring, please also answer two pairwise questions: (i) is the Top-8 KG hypothesis better than the No-KG hypothesis, worse than it, or about the same? (ii) is the Full KG hypothesis better than the No-KG hypothesis, worse than it, or about the same?

Please do not reward length alone. Prefer hypotheses that clearly connect the problem, intervention, mechanism, and target property in a scientifically plausible way.
\end{quote}

We report the mean scores across the five examples in Table~\ref{tab:expert_case_study}. Because this assessment uses one rater and five examples, we interpret it only as a qualitative check that graph-conditioned outputs are more mechanistically grounded in representative cases.

\subsection{Human expert Case Study Examples}
\label{app:domain_rater_cases}

\begin{tcolorbox}[
colback=white!4,
colframe=blue!65,
title=\textbf{Five case-study examples used for the domain-rater check},
fonttitle=\bfseries,
breakable,
left=1mm,
right=1mm,
top=1mm,
bottom=1mm
]
\small
We selected five examples where the full-KG output recovered graph content that the no-KG output missed. In all five cases, the full-KG output had TRR = 1.0, the no-KG output had TRR = 0.0, and $\Delta$TRR = 1.0. The goal was not to run a large human evaluation, but to check whether higher graph recovery corresponded to qualitatively better scientific hypotheses.

\begin{enumerate}
    \item \textbf{Paper 1}: Thermal management for lithium-ion UUV batteries. The full-KG output specifies a pressure-compensating phase-change material system and links it to thermal instability, capacity fade, and impedance growth.
    
    \item \textbf{Paper 2}: Additive manufacturing of sodium-ion battery components. The full-KG output focuses on thixotropic, shear-thinning inks and volatile co-solvent control to reduce coffee-ring effects and non-uniform morphology.
    
    \item \textbf{Paper 3}: Polymer-gel electrolyte design. The full-KG output specifies a low-molecular-weight fluorinated ether additive and links it to dense electrical double-layer ordering, screening, and desolvation impedance.
    
    \item \textbf{Paper 4}: High-nickel NCM cathode stabilization. The full-KG output proposes a conformal LiAlO$_2$ coating and connects it to hydrofluoric-acid neutralization, parasitic surface reactions, and cycling stability.
    
    \item \textbf{Paper 5}: Lithium-ion battery recycling. The full-KG output proposes mild electrochemical dissolution at the cathode-current collector interface to weaken binder adhesion and reduce material loss and cross-contamination.
\end{enumerate}
\end{tcolorbox}


\subsection{Model-Specific Top Graph Examples and Prompting Guidance}
\label{app:top_graph_examples}

Table~\ref{tab:model_guidance_appendix} summarizes the model-specific prompting guidance derived from the best-performing graph conditions. The representative examples in Box~\ref{box:top_graph_examples} show that the strongest graph prompts preserve a materials-science chain from failure mode to intervention, mechanism, target property, and outcome.

\begin{table*}[t]
\centering
\small
\setlength{\tabcolsep}{4pt}
\caption{
\textbf{Model-specific graph prompting guidance.}
The most effective graph style differs by model, but useful prompts preserve the materials-science chain from failure mode to intervention, mechanism, and target property.
}
\label{tab:model_guidance_appendix}
\begin{tabular}{p{0.14\linewidth}p{0.26\linewidth}p{0.50\linewidth}}
\toprule
Model & Best graph style & Practical guidance \\
\midrule
Gemini 
& Sparse, rich, full-path 
& Use concise typed chains linking failure, intervention, mechanism, target property, and outcome. Avoid large undifferentiated graph dumps. \\
Llama-3.1-70B 
& Sparse, coarse, 2-hop 
& Use short local graphs centered on the problem, with simple relation labels. Concise local grounding appears sufficient. \\
Mistral-7B 
& Dense, rich, 2-hop 
& Provide more explicit local scaffolding with clear relation labels. Avoid noisy or shuffled relations. \\
\bottomrule
\end{tabular}
\end{table*}

\begin{tcolorbox}[
colback=gray!4,
colframe=blue!65,
title=\textbf{Representative top graph prompts by model},
fonttitle=\bfseries,
breakable,
left=1mm,
right=1mm,
top=1mm,
bottom=1mm,
label={box:top_graph_examples}
]
\small

\textbf{Gemini: Sparse + T3 multihop + full path.}
The selected example concerns metal--iodine batteries, where rapid capacity degradation and self-discharge arise from polyiodide dissolution and migration. The graph-conditioned output achieves RFS$_{\mathrm{ref}}=1.000$, TRR$_{\mathrm{ref}}=0.750$, and an RFS gain of $+0.800$ over no KG.

\textbf{Materials-science chain in the supplied graph:}
\begin{itemize}
    \item Failure mode $\rightarrow$ polyiodide shuttle effect and low electronic conductivity
    \item Intervention $\rightarrow$ N,P-doped hierarchical porous carbon hosts with quaternary ammonium coating
    \item Mechanism $\rightarrow$ synergistic chemisorption by heteroatoms and physical confinement
    \item Target property $\rightarrow$ cycle life and rate capability
    \item Outcome $\rightarrow$ suppressed self-discharge and enhanced electrochemical stability
\end{itemize}

\textbf{Prompting takeaway:} Gemini benefits from compact, semantically rich paths that connect failure, intervention, mechanism, and outcome.

\medskip

\textbf{Llama-3.1-70B: Sparse + T1 coarse + 2-hop.}
The selected example concerns high-capacity lithium-ion anodes with volumetric expansion and unstable SEI formation. The graph-conditioned output achieves RFS$_{\mathrm{ref}}=1.000$, TRR$_{\mathrm{ref}}=0.375$, and an RFS gain of $+0.600$ over no KG.

\textbf{Materials-science chain in the supplied graph:}
\begin{itemize}
    \item Failure mode $\rightarrow$ mechanical pulverization and continuous electrolyte decomposition
    \item Intervention $\rightarrow$ polydopamine-derived nitrogen-doped carbon nanoshell overcoating
    \item Mechanism $\rightarrow$ mechanically resilient and ionically conductive barrier
    \item Target property $\rightarrow$ cycling stability and rate performance
    \item Outcome $\rightarrow$ enhanced capacity retention and prevention of active-material pulverization
\end{itemize}

\textbf{Prompting takeaway:} Llama-3.1-70B benefits from concise local graphs with simple relation labels.

\medskip

\textbf{Mistral-7B: Dense + T3 multihop + 2-hop.}
The selected example again concerns metal-iodine batteries with the polyiodide shuttle effect. The graph-conditioned output achieves RFS$_{\mathrm{ref}}=0.600$, TRR$_{\mathrm{ref}}=0.875$, and an RFS gain of $+0.400$ over no KG.

\textbf{Materials-science chain in the supplied graph:}
\begin{itemize}
    \item Failure mode $\rightarrow$ polyiodide shuttle effect and low electronic conductivity
    \item Intervention $\rightarrow$ N,P-doped hierarchical porous carbon hosts with quaternary ammonium coating
    \item Mechanism $\rightarrow$ synergistic chemisorption by heteroatoms and physical confinement
    \item Target property $\rightarrow$ cycle life and rate capability
    \item Outcome $\rightarrow$ suppressed self-discharge and enhanced electrochemical stability
\end{itemize}

\textbf{Prompting takeaway:} Mistral-7B benefits from denser local scaffolding with explicit relation labels.
\end{tcolorbox}

\begin{table*}[t]
\centering
\small
\setlength{\tabcolsep}{4pt}
\caption{
\textbf{Qualitative example.}
The full-KG condition makes the hypothesis more specific by grounding the intervention in a mechanism involving electrical double layer screening and desolvation impedance.
}
\label{tab:case_study}
\begin{tabularx}{\textwidth}{lX}
\toprule
Condition & Output excerpt \\
\midrule
No KG & The incorporation of tailored amphiphilic additives into a polymer-gel electrolyte will lower the Li-ion desolvation energy barrier and reduce interfacial resistance by disrupting the dense electrical double layer's screening effect. \\
Full KG & The incorporation of a low-molecular-weight fluorinated ether additive into the polymer-gel electrolyte will disrupt dense electrical double layer ordering and reduce its screening effect, thereby mitigating electrical double layer screening and desolvation impedance. \\
Observation & The no-KG output is plausible but generic. The full-KG output specifies the additive class and ties it more directly to the failure mechanism and target impedance pathway. \\
\bottomrule
\end{tabularx}
\end{table*}

\section{Metric Implementation and KG Controls}
\label{app:implementation}

\paragraph{Data accounting.}
The core dataset contains 100 unique battery-science problems. Larger row counts in the appendix arise from aggregation over models, KG conditions, top-$k$ values, or repeated samples. For example, the main pipeline contains 100 problems $\times$ 11 KG conditions $\times$ 3 models = 3,300 generations. The contradiction experiment contains 50 problems $\times$ 9 conditions $\times$ 3 models = 1,350 generations. The sufficiency experiment contains 100 problems $\times$ 12 graph settings $\times$ 3 models = 3,600 generations. The centrality analysis contains 30 problems $\times$ 4 values of $k$ $\times$ 5 selection methods $\times$ 2 models = 1,200 generations. We specify the aggregation unit in each table caption. 

\paragraph{Compute and infrastructure.}
Open-model inference was run on H200 GPUs and Gemini 2.5 Flash was accessed through an API. We do not train or fine-tune models. The exact provider-side compute and parameter counts for Gemini 2.5 Flash are not publicly available.


\subsection{Computing RFS and KTC}
\label{app:rfs_ktc}

Relation Fidelity Score (RFS) is a deterministic, rule-based metric that measures whether relation roles present in the supplied KG are reflected in the generated hypothesis. Each KG edge is typed when the graph is built from structured scientific fields. We map typed edges into broad scientific roles, including failure, intervention, mechanism, property, component, system, and outcome. For each relation role present in the KG context, we check whether the generated hypothesis contains language associated with that role.

Each role is associated with a small manually defined keyword inventory. For example, mechanism relations are associated with cues such as \emph{because}, \emph{via}, \emph{through}, \emph{mechanism}, and \emph{mediated}; failure relations with cues such as \emph{degradation}, \emph{capacity fade}, \emph{instability}, \emph{dissolution}, and \emph{plating}; intervention relations with cues such as \emph{coating}, \emph{doping}, \emph{modification}, \emph{engineering}, and \emph{treatment}; and outcome/property relations with cues such as \emph{improve}, \emph{enhance}, \emph{reduce}, \emph{suppress}, and \emph{achieve}. RFS is computed as:
\[
\mathrm{RFS}(y,G)=
\frac{1}{|R(G)|}
\sum_{r \in R(G)}
\mathbf{1}[\mathrm{signals}(r) \cap y \neq \emptyset],
\]
where $R(G)$ is the set of relation roles in the supplied KG and $y$ is the generated hypothesis. Because RFS is rule-based and deterministic, inter-annotator agreement is not applicable; we interpret it as an operational diagnostic of relation-language preservation, not as an expert semantic parser.

KG Triple Coverage (KTC) measures how much of the KG object-side content is reflected in the generated hypothesis. For each KG triple supplied $(s_i,r_i,o_i)$, we extract normalized content words from the object $o_i$ and compare them to normalized content words in the generated hypothesis:
\[
\mathrm{KTC}(y,G)=
\frac{|\mathrm{terms}(O_G) \cap \mathrm{terms}(y)|}
{|\mathrm{terms}(O_G)|},
\]
where $O_G$ is the set of object strings in the supplied KG. KTC differs from TRR in granularity: TRR asks whether a triple's object entity is recalled, while KTC measures broader object-side term coverage.

\paragraph{RFS/KTC reliability.} RFS and KTC are deterministic diagnostic metrics, not human annotation metrics. RFS uses a fixed relation-role keyword inventory to test whether relation-role language appears in the hypothesis. KTC measures the overlap of the normalized content on the object-side between the supplied KG and the output. Because the labels are produced by fixed rules, inter-annotator agreement is not applicable. We interpret these metrics as proxies for graph use rather than expert judgments of scientific quality.

\paragraph{Degenerate graph conditions.}
TRR, RFS, and KTC are defined relative to the graph information supplied in a condition. In the no-KG condition, no KG objects or relations are provided, so graph-use metrics are set to 0 by convention and interpreted as ``no graph signal available,'' not as a general judgment of output quality. Similarly, in the entity-only condition, object strings are provided but relation labels are removed; RFS is therefore 0 because there are no supplied relation roles to preserve. In the relation-skeleton condition, relation roles are provided but concrete object entities are masked, so RFS can be nonzero while TRR and KTC remain near zero. This behavior is intentional: the metrics diagnose which kind of graph information is available and used.

\subsection{Top-$k$ Triple Ranking}
\label{app:topk_ranking}

The targeted and compressed KG variants use a deterministic relevance score to rank  triples before selecting the top subset $k$. For each edge, we score the object label against the problem statement:
\[
s(e)=
\frac{|\mathrm{words}(o_e) \cap \mathrm{words}(p)|}
{|\mathrm{words}(o_e)|}
\times b(e),
\]
where $o_e$ is the object label for edge $e$, $p$ is the problem statement, and $b(e)$ is a relation-type boost. We set $b(e)=1.3$ for mechanism, failure, and intervention triples, because these roles are central to scientific hypothesis generation, and $b(e)=1.0$ otherwise. This ranking is intentionally simple and transparent; it is used to test whether a small relevance-ranked subset of graph facts can recover full-KG behavior.

\begin{table*}[t]
\centering
\small
\setlength{\tabcolsep}{4pt}
\caption{
\textbf{Necessity-control results across KG variants.}
Entity-only context does not recover graph-grounded behavior. Relation skeletons recover relation-fidelity signal but not entity or coverage metrics. Targeted KG recovers much of the full-KG behavior, supporting the claim that useful graph signal is concentrated in a compact subset.
}
\label{tab:kg_variant_results}
\resizebox{0.7\textwidth}{!}{
\begin{tabular}{lccccccccc}
\toprule
Variant
& \multicolumn{3}{c}{Gemini}
& \multicolumn{3}{c}{Llama-3.1-70B}
& \multicolumn{3}{c}{Mistral-7B} \\
\cmidrule(lr){2-4}\cmidrule(lr){5-7}\cmidrule(lr){8-10}
& TRR & RFS & KTC & TRR & RFS & KTC & TRR & RFS & KTC \\
\midrule
No KG         & 0.000 & 0.000 & 0.000 & 0.000 & 0.000 & 0.000 & 0.000 & 0.000 & 0.000 \\
Entity-only   & 0.000 & 0.000 & 0.000 & 0.000 & 0.000 & 0.000 & 0.000 & 0.000 & 0.000 \\
Rel. skeleton & 0.000 & 0.544 & 0.000 & 0.010 & 0.570 & 0.000 & 0.045 & 0.574 & 0.000 \\
Targeted KG   & 0.780 & 0.640 & 0.598 & 0.730 & 0.629 & 0.546 & 0.855 & 0.600 & 0.664 \\
Full KG       & 0.594 & 0.546 & 0.526 & 0.661 & 0.546 & 0.541 & 0.799 & 0.596 & 0.640 \\
\bottomrule
\end{tabular}}
\end{table*}

\begin{table*}[t]
\centering
\small
\setlength{\tabcolsep}{4pt}
\caption{
\textbf{Context-length distribution by KG variant.}
We report the distribution of KG context lengths used in the necessity-control experiment. These lengths help check whether gains are driven simply by longer prompts.
}
\label{tab:kg_context_lengths}
\resizebox{0.7\textwidth}{!}{
\begin{tabular}{lrrrrrrrr}
\toprule
Variant & Count & Mean & Std. & Min & 25\% & 50\% & 75\% & Max \\
\midrule
Entity-only       & 200 & 648.1  & 36.6 & 555.0  & 623.5  & 650.5  & 669.0  & 745.0 \\
Full KG           & 200 & 1348.1 & 36.6 & 1255.0 & 1323.5 & 1350.5 & 1369.0 & 1445.0 \\
No KG             & 200 & 64.0   & 0.0  & 64.0   & 64.0   & 64.0   & 64.0   & 64.0 \\
Relation skeleton & 200 & 664.6  & 7.9  & 644.0  & 659.0  & 664.5  & 670.2  & 685.0 \\
Targeted KG       & 200 & 1080.3 & 40.2 & 999.0  & 1053.5 & 1085.0 & 1105.0 & 1197.0 \\
\bottomrule
\end{tabular}}
\end{table*}
\subsection{Random, Shuffled, Entity-Only, and Relation-Skeleton Controls}
\label{app:kg_controls}

We use several KG variants to separate different possible sources of graph utility. The random KG condition replaces the problem's real KG with a KG subgraph sampled from a different problem. Where possible, we sample a random subgraph with the same number of triples as the corresponding real condition. This controls for the number of explicit facts, although exact token length may still differ because object labels vary in length.

The shuffled KG condition keeps the entities from the real KG but randomly permutes the relation labels. This preserves much of the surface content while breaking the mapping between entities and their scientific roles. The comparison between real KG and shuffled KG therefore tests whether the model is sensitive to relation structure rather than only entity co-occurrence.

The entity-only condition removes relation labels and keeps only object-side entity strings. This tests whether graph gains can be explained by lexical exposure to scientific terms. The relation-skeleton condition preserves relation types and typed placeholders while masking concrete entity identity. This tests whether relation-role information alone induces relation-consistent language.

\subsection{Necessity-Control Results}
\label{app:necessity_controls}

We next ask what kind of graph information is actually necessary for the observed gains. To test this, we compare the full KG against four reduced variants: no KG, entity-only context, relation skeleton, targeted KG, and full KG. This experiment isolates whether performance depends on raw entity strings, relation type information, or a compact targeted subset of graph facts.

The results show that, entity-only context is not enough. Across models, providing only entity names produces near-zero performance on the main KG-grounding metrics, indicating that simple lexical exposure to graph entities does not recover the full-KG effect. Additionally, relation skeleton preserves mostly relation fidelity. It recovers some RFS while leaving TRR and KTC near zero. This shows that relation-type information is meaningful on its own, but it is not sufficient to reproduce the full behavior of graph-conditioned hypothesis generation. Third, targeted KG recovers most of the benefit of the full graph and in several cases matches or slightly exceeds the full-KG condition. Table~\ref{tab:kg_variant_results} reports these results.

\subsection{Context Length Distributions}
\label{app:context_lengths}

Table~\ref{tab:kg_context_lengths} reports the distribution of KG context lengths for the necessity variants. These values are used to check whether the intervention results could be explained only by prompt length. The variants differ in length because they intentionally remove or preserve different kinds of information. Full KG is longest, targeted KG is shorter but still contains concrete entity and relation content, while entity-only and relation-skeleton variants are much shorter. No KG is a fixed short baseline.

These length differences are important because long-context models may not use all parts of the context uniformly. Prior work on long-context language models shows that performance can depend on where relevant information appears in the input, with models often struggling to use information placed in the middle of long contexts. We therefore interpret the length-controlled controls, especially random and shuffled KG, together with the targeted and top-$k$ experiments.

\subsection{KG Variant Definitions}
\label{app:kg_variant_definitions}

All graph conditions are constructed from a per-problem local KG $G_p=(V_p,E_p)$. Each edge $e=(s,r,o)\in E_p$ is a typed subject--relation--object triple derived from structured scientific fields such as material system, component, failure mode, intervention, mechanism, target property, and claimed outcome. The full local KG contains 15--18 triples per problem.

\begin{table*}[t]
\centering
\small
\setlength{\tabcolsep}{4pt}
\caption{
\textbf{Definitions of KG variants.}
All variants are generated from the same per-problem local KG. Density, ontology, and topology variants change the amount or form of graph context, while control variants test whether models use relevant graph content and relation structure.
}
\label{tab:kg_variant_definitions}
\resizebox{\textwidth}{!}{
\begin{tabular}{lll}
\toprule
Variant group & Condition & Definition \\
\midrule
No graph & No KG & Prompt contains the scientific problem but no graph triples. \\

Density & Sparse & Small relevance-ranked subset of $E_p$ with few graph facts. \\
Density & Medium & Intermediate subset of $E_p$ with more graph facts than sparse. \\
Density & Dense & Expanded verbalization of the local graph with the largest graph context. \\

Ontology & T1 coarse & Coarse relation labels over the same problem fields. \\
Ontology & T3 multihop & Richer relation labels and multihop scientific roles. \\

Topology & 2-hop & Triples within a local two-hop neighborhood of the problem node. \\
Topology & Full path & Longer problem-to-field/path view retaining extended graph paths. \\

Controls & Random KG & Same triple count as the matched real condition, but sampled from another problem. \\
Controls & Shuffled KG & Same entities as the real KG, but relation labels are randomly permuted. \\
Controls & Entity-only & Keeps object-side entity strings but removes relation labels. \\
Controls & Relation skeleton & Keeps relation types/placeholders but masks concrete entity identity. \\

Compression & Targeted KG & Relevance-ranked subset selected by the deterministic score in Appendix~\ref{app:topk_ranking}. \\
Compression & Top-$k$ KG & Keeps the top $k$ ranked triples, with $k\in\{1,2,4,8\}$. \\
\bottomrule
\end{tabular}}
\end{table*}

\begin{table*}[t]
\centering
\small
\setlength{\tabcolsep}{4.5pt}
\caption{Paired permutation tests for the two most important control comparisons: replacing the real KG with a random KG, and shuffling KG structure while keeping content. Positive $\Delta$ indicates that the first condition outperforms the second. Significance: $^{*}p<0.05$, $^{**}p<0.01$, $^{***}p<0.001$.}
\label{tab:perm_controls}
\resizebox{0.6\textwidth}{!}{
\begin{tabular}{lllrcl}
\toprule
Model & Metric & Comparison & $\Delta$ & $p$-value & Sig. \\
\midrule
Gemini 2.5 Flash        & TRR  & Real KG vs Random KG   & +0.2900 & 0.0000 & $^{***}$ \\
Gemini 2.5 Flash        & RFS  & Real KG vs Random KG   & +0.0033 & 0.8896 & ns \\
Gemini 2.5 Flash        & KTC  & Real KG vs Random KG   & +0.2559 & 0.0000 & $^{***}$ \\
Gemini 2.5 Flash        & FGS  & Real KG vs Random KG   & +0.0272 & 0.0574 & ns \\
Gemini 2.5 Flash        & CHAS & Real KG vs Random KG   & +0.0685 & 0.0064 & $^{**}$ \\
Mistral-7B    & TRR  & Real KG vs Random KG   & +0.0080 & 1.0000 & ns \\
Mistral-7B    & RFS  & Real KG vs Random KG   & +0.0067 & 1.0000 & ns \\
Mistral-7B    & KTC  & Real KG vs Random KG   & +0.0054 & 1.0000 & ns \\
Mistral-7B    & FGS  & Real KG vs Random KG   & +0.0061 & 1.0000 & ns \\
Mistral-7B    & CHAS & Real KG vs Random KG   & +0.0062 & 1.0000 & ns \\
Llama-3.1-70B & TRR  & Real KG vs Random KG   & +0.0380 & 0.0708 & ns \\
Llama-3.1-70B & RFS  & Real KG vs Random KG   & +0.0133 & 0.5530 & ns \\
Llama-3.1-70B & KTC  & Real KG vs Random KG   & +0.0331 & 0.0426 & $^{*}$ \\
Llama-3.1-70B & FGS  & Real KG vs Random KG   & -0.0043 & 0.8570 & ns \\
Llama-3.1-70B & CHAS & Real KG vs Random KG   & +0.0076 & 0.7648 & ns \\
\midrule
Gemini 2.5 Flash        & TRR  & Real KG vs Shuffled KG & -0.0380 & 0.0198 & $^{*}$ \\
Gemini 2.5 Flash        & RFS  & Real KG vs Shuffled KG & +0.1867 & 0.0000 & $^{***}$ \\
Gemini 2.5 Flash       & KTC  & Real KG vs Shuffled KG & -0.0183 & 0.1842 & ns \\
Gemini 2.5 Flash       & FGS  & Real KG vs Shuffled KG & -0.0087 & 0.4980 & ns \\
Gemini 2.5 Flash        & CHAS & Real KG vs Shuffled KG & -0.0304 & 0.1842 & ns \\
Mistral-7B    & TRR  & Real KG vs Shuffled KG & -0.0900 & 0.0000 & $^{***}$ \\
Mistral-7B    & RFS  & Real KG vs Shuffled KG & -0.0800 & 0.0000 & $^{***}$ \\
Mistral-7B    & KTC  & Real KG vs Shuffled KG & -0.0731 & 0.0000 & $^{***}$ \\
Mistral-7B    & FGS  & Real KG vs Shuffled KG & -0.0883 & 0.0000 & $^{***}$ \\
Mistral-7B    & CHAS & Real KG vs Shuffled KG & -0.0827 & 0.0000 & $^{***}$ \\
Llama-3.1-70B & TRR  & Real KG vs Shuffled KG & -0.0240 & 0.3656 & ns \\
Llama-3.1-70B & RFS  & Real KG vs Shuffled KG & +0.0100 & 0.7340 & ns \\
Llama-3.1-70B & KTC  & Real KG vs Shuffled KG & -0.0151 & 0.4534 & ns \\
Llama-3.1-70B & FGS  & Real KG vs Shuffled KG & -0.0106 & 0.5746 & ns \\
Llama-3.1-70B & CHAS & Real KG vs Shuffled KG & -0.0113 & 0.6782 & ns \\
\bottomrule
\end{tabular}
}
\end{table*}

\section{Additional Statistical Results}
\label{app:stats}

\begin{table}[t]
\centering
\small
\setlength{\tabcolsep}{5pt}
\caption{Mixed-effects variance decomposition across evaluation metrics. For all three metrics, model identity explains more variance than condition identity, indicating that performance differences are driven more strongly by the underlying LLM than by KG condition alone.}
\label{tab:mixed_effects}
\begin{tabular}{lccc}
\toprule
Metric & Dominant factor & Model effect & Condition effect \\
\midrule
TRR  & model & 0.1632 & 0.0288 \\
RFS  & model & 0.2095 & 0.0928 \\
KTC  & model & 0.2254 & 0.1358 \\
\bottomrule
\end{tabular}
\end{table}

\subsection{Paired Permutation Tests}
\label{app:permutation_tests}

We use paired permutation tests for condition-level contrasts. For each problem, we compute the metric difference between two KG conditions and test whether the mean paired difference is larger or smaller than expected under random sign flips. This is a non-parametric repeated-measures test and does not assume normality of metric values.

\subsubsection{Control Comparisons}
\label{app:control_comparisons}

Table~\ref{tab:perm_controls} reports the two main control comparisons. The real-vs-random comparison tests whether models benefit from relevant graph facts rather than generic extra context. The real-vs-shuffled comparison keeps much of the surface content but corrupts relation structure, so it tests whether models are sensitive to the organization of the graph. These results support the main-text claim that Gemini shows the strongest and most stable use of KG structure, Llama-3.1-70B shows smaller gains, and Mistral-7B is weaker and more brittle under corrupted graph structure.

\subsubsection{Structural Ablations}
\label{app:structural_ablations}

Table~\ref{tab:perm_ablations} reports paired tests for the structural KG ablations. These comparisons test whether graph density, ontology richness, and topology change how much KG information appears in the generated output. The sparse-vs-medium comparison evaluates whether adding more graph facts helps or hurts. The T1-vs-T3 comparison tests coarse versus richer ontology structure. The 2-hop-vs-full-path comparison tests whether deeper topology adds useful context beyond the shorter local graph. These results support the main-text claim that more graph context is not uniformly better.

\begin{table*}[t]
\centering
\small
\setlength{\tabcolsep}{4.5pt}
\caption{Paired permutation tests for KG ablations. Positive $\Delta$ indicates that the first condition outperforms the second. Significance: $^{*}p<0.05$, $^{**}p<0.01$, $^{***}p<0.001$.}
\label{tab:perm_ablations}
\resizebox{0.6\textwidth}{!}{
\begin{tabular}{lllrcl}
\toprule
Model & Metric & Comparison & $\Delta$ & $p$-value & Sig. \\
\midrule
Gemini        & TRR  & Sparse vs Medium density & +0.1010 & 0.0000 & $^{***}$ \\
Gemini        & RFS  & Sparse vs Medium density & +0.0100 & 0.7052 & ns \\
Gemini        & KTC  & Sparse vs Medium density & +0.1440 & 0.0000 & $^{***}$ \\
Gemini        & FGS  & Sparse vs Medium density & +0.0091 & 0.4162 & ns \\
Gemini        & CHAS & Sparse vs Medium density & -0.0346 & 0.1490 & ns \\
Mistral-7B    & TRR  & Sparse vs Medium density & +0.0420 & 0.0302 & $^{*}$ \\
Mistral-7B    & RFS  & Sparse vs Medium density & +0.0483 & 0.0302 & $^{*}$ \\
Mistral-7B    & KTC  & Sparse vs Medium density & +0.0446 & 0.0302 & $^{*}$ \\
Mistral-7B    & FGS  & Sparse vs Medium density & +0.0394 & 0.0302 & $^{*}$ \\
Mistral-7B    & CHAS & Sparse vs Medium density & +0.0430 & 0.0302 & $^{*}$ \\
Llama-3.1-70B & TRR  & Sparse vs Medium density & +0.0420 & 0.0940 & ns \\
Llama-3.1-70B & RFS  & Sparse vs Medium density & +0.0317 & 0.1486 & ns \\
Llama-3.1-70B & KTC  & Sparse vs Medium density & +0.0475 & 0.0336 & $^{*}$ \\
Llama-3.1-70B & FGS  & Sparse vs Medium density & +0.0138 & 0.4814 & ns \\
Llama-3.1-70B & CHAS & Sparse vs Medium density & +0.0243 & 0.2840 & ns \\
\midrule
Gemini        & TRR  & T1-coarse vs T3-multihop & -0.0080 & 0.6318 & ns \\
Gemini        & RFS  & T1-coarse vs T3-multihop & +0.0067 & 0.7676 & ns \\
Gemini        & KTC  & T1-coarse vs T3-multihop & +0.0012 & 0.9316 & ns \\
Gemini        & FGS  & T1-coarse vs T3-multihop & +0.0076 & 0.4336 & ns \\
Gemini        & CHAS & T1-coarse vs T3-multihop & -0.0168 & 0.3218 & ns \\
Mistral-7B    & TRR  & T1-coarse vs T3-multihop & -0.0080 & 1.0000 & ns \\
Mistral-7B    & RFS  & T1-coarse vs T3-multihop & -0.0067 & 1.0000 & ns \\
Mistral-7B    & KTC  & T1-coarse vs T3-multihop & -0.0038 & 1.0000 & ns \\
Mistral-7B    & FGS  & T1-coarse vs T3-multihop & -0.0008 & 1.0000 & ns \\
Mistral-7B    & CHAS & T1-coarse vs T3-multihop & +0.0006 & 1.0000 & ns \\
Llama-3.1-70B & TRR  & T1-coarse vs T3-multihop & +0.0340 & 0.1838 & ns \\
Llama-3.1-70B & RFS  & T1-coarse vs T3-multihop & +0.0100 & 0.7106 & ns \\
Llama-3.1-70B & KTC  & T1-coarse vs T3-multihop & +0.0234 & 0.2126 & ns \\
Llama-3.1-70B & FGS  & T1-coarse vs T3-multihop & +0.0268 & 0.2298 & ns \\
Llama-3.1-70B & CHAS & T1-coarse vs T3-multihop & +0.0290 & 0.2348 & ns \\
\midrule
Gemini        & TRR  & 2-hop vs Full-path & -0.0320 & 0.0346 & $^{*}$ \\
Gemini        & RFS  & 2-hop vs Full-path & -0.0200 & 0.3746 & ns \\
Gemini        & KTC  & 2-hop vs Full-path & -0.0259 & 0.0416 & $^{*}$ \\
Gemini        & FGS  & 2-hop vs Full-path & -0.0003 & 0.9836 & ns \\
Gemini        & CHAS & 2-hop vs Full-path & +0.0044 & 0.7836 & ns \\
Mistral-7B    & TRR  & 2-hop vs Full-path & +0.0100 & 0.5098 & ns \\
Mistral-7B    & RFS  & 2-hop vs Full-path & +0.0100 & 0.7540 & ns \\
Mistral-7B    & KTC  & 2-hop vs Full-path & +0.0072 & 0.7540 & ns \\
Mistral-7B    & FGS  & 2-hop vs Full-path & +0.0073 & 0.5098 & ns \\
Mistral-7B    & CHAS & 2-hop vs Full-path & +0.0093 & 0.5098 & ns \\
Llama-3.1-70B & TRR  & 2-hop vs Full-path & +0.0140 & 0.4312 & ns \\
Llama-3.1-70B & RFS  & 2-hop vs Full-path & +0.0233 & 0.2032 & ns \\
Llama-3.1-70B & KTC  & 2-hop vs Full-path & +0.0150 & 0.3534 & ns \\
Llama-3.1-70B & FGS  & 2-hop vs Full-path & +0.0132 & 0.5216 & ns \\
Llama-3.1-70B & CHAS & 2-hop vs Full-path & +0.0270 & 0.2604 & ns \\
\bottomrule
\end{tabular}
}
\end{table*}

\subsection{Bootstrap Confidence Intervals}
\label{app:bootstrap}

Bootstrap confidence intervals provide uncertainty estimates for the main effect sizes reported in the paper. Unlike the permutation tests, which test whether a paired difference is unlikely under a null hypothesis, the bootstrap intervals estimate the likely range of each effect size. We use them to check whether the main directional effects are stable across resampled problem sets. Intervals that do not cross zero indicate stable directional effects.

Table~\ref{tab:bootstrap_ci} supports the main claims in the paper. Gemini 2.5 Flash has stable positive gains for real KG over random KG, real KG over no KG, and real KG over shuffled KG on relation fidelity. Llama-3.1-70B shows smaller positive effects, especially for KTC. Mistral-7B shows weak real-vs-random gains and a negative real-vs-shuffled RFS effect, consistent with brittle use of relation structure.

\begin{table*}[t]
\centering
\small
\setlength{\tabcolsep}{5pt}
\caption{
\textbf{Bootstrap 95\% confidence intervals for the main KG-effect contrasts.}
Intervals that do not cross zero indicate stable directional effects. These intervals support the main-text claim that Gemini 2.5 Flash has the strongest and most stable KG gains, Llama-3.1-70B shows smaller positive gains, and Mistral-7B remains weak or brittle under several graph manipulations.
}
\label{tab:bootstrap_ci}
\begin{tabular}{llccc}
\toprule
Model & Quantity & Estimate & 95\% CI lower & 95\% CI upper \\
\midrule
Gemini 2.5 Flash        & $\Delta$TRR(real$-$random)   & +0.2900 & +0.2420 & +0.3420 \\
Mistral-7B    & $\Delta$TRR(real$-$random)   & +0.0080 & +0.0000 & +0.0240 \\
Llama-3.1-70B & $\Delta$TRR(real$-$random)   & +0.0380 & +0.0000 & +0.0800 \\
\midrule
Gemini 2.5 Flash        & $\Delta$RFS(real$-$shuffled) & +0.1867 & +0.1333 & +0.2433 \\
Mistral-7B    & $\Delta$RFS(real$-$shuffled) & -0.0800 & -0.1267 & -0.0367 \\
Llama-3.1-70B & $\Delta$RFS(real$-$shuffled) & +0.0100 & -0.0333 & +0.0533 \\
\midrule
Gemini 2.5 Flash       & $\Delta$KTC(real$-$noKG)     & +0.7569 & +0.7305 & +0.7832 \\
Mistral-7B    & $\Delta$KTC(real$-$noKG)     & +0.0054 & +0.0000 & +0.0162 \\
Llama-3.1-70B & $\Delta$KTC(real$-$noKG)     & +0.0621 & +0.0287 & +0.0998 \\
\midrule
Gemini 2.5 Flash       & $\Delta$sparse$-$dense       & +0.1170 & +0.0760 & +0.1535 \\
Mistral-7B    & $\Delta$sparse$-$dense       & -0.0389 & -0.0879 & +0.0092 \\
Llama-3.1-70B & $\Delta$sparse$-$dense       & +0.0476 & -0.0127 & +0.1075 \\
\midrule
Gemini 2.5 Flash        & $\Delta$T1$-$T3              & -0.0080 & -0.0380 & +0.0240 \\
Mistral-7B    & $\Delta$T1$-$T3              & -0.0080 & -0.0300 & +0.0040 \\
Llama-3.1-70B & $\Delta$T1$-$T3              & +0.0340 & -0.0120 & +0.0800 \\
\midrule
Gemini 2.5 Flash       & $\Delta$2hop$-$fullpath      & -0.0320 & -0.0620 & -0.0020 \\
Mistral-7B    & $\Delta$2hop$-$fullpath      & +0.0100 & -0.0080 & +0.0340 \\
Llama-3.1-70B & $\Delta$2hop$-$fullpath      & +0.0140 & -0.0200 & +0.0500 \\
\bottomrule
\end{tabular}
\end{table*}

\section{Robustness Checks}
\label{app:robustness}

We include two additional robustness checks. First, we compare our semantic top-$k$ ranking against matched random and topology-based top-$k$ selectors. This tests whether the compression result is driven only by semantic relevance, by shorter prompts, or by graph centrality. Second, we estimate sampling variability by generating multiple outputs for a subset of conditions.

\subsection{Top-$k$ Semantic Ranking Versus Topology Baselines}
\label{app:ranking_baselines}

The main experiments use a simple semantic relevance score to select top-$k$ triples. To test whether the compression effect depends on this particular ranking rule, we compare semantic top-$k$ selection against four matched baselines: random-$k$, degree-$k$, betweenness-$k$, and PageRank-$k$. All methods select the same number of triples, so the comparison controls for the size of the compressed graph. Degree, betweenness, and PageRank provide topology-based alternatives to the semantic ranking.

Table~\ref{tab:ranking_baselines} reports semantic distance to the full-KG output. Lower values mean that the selected top-$k$ subset better approximates the behavior induced by the full graph. We observe that compression is robust. For both Llama-3.1-70B and Mistral-7B, increasing $k$ consistently reduces semantic distance for all ranking methods. However, semantic ranking is not always the best selector. Random and topology-based methods sometimes match or outperform the semantic ranking at particular values of $k$. We therefore do not claim that our semantic ranking is optimal. Instead, these results show that the central compression effect is not an artifact of one ranking heuristic: several small matched subsets can recover much of the full-KG behavior.

We also recompute the ranking-baseline comparison using fixed-reference TRR. This avoids the smaller-denominator issue for top-$k$ subsets. The result reinforces the main interpretation: compression is robust, but not selector-unique. Random and topology-based subsets can match or exceed semantic top-$k$ under fixed-reference scoring, especially when $k=8$.

\begin{table*}[t]
\centering
\small
\setlength{\tabcolsep}{5pt}
\caption{
\textbf{Fixed-reference TRR for top-$k$ selection methods.}
All methods are scored against the same full KG. Random and topology-based selectors sometimes match or exceed semantic top-$k$, especially at larger $k$. This supports the compression claim while showing that compression is not unique to one semantic ranking rule.
}
\label{tab:centrality_fixed_reference}
\begin{tabular}{llccccc}
\toprule
Model & $k$ & Betweenness & Degree & PageRank & Random & Semantic \\
\midrule
\multirow{4}{*}{Llama-70B}
& 1 & 0.396 & 0.381 & 0.384 & 0.441 & 0.396 \\
& 2 & 0.431 & 0.442 & 0.453 & 0.506 & 0.419 \\
& 4 & 0.512 & 0.514 & 0.537 & 0.608 & 0.520 \\
& 8 & 0.774 & 0.773 & 0.777 & 0.801 & 0.771 \\
\midrule
\multirow{4}{*}{Mistral-7B}
& 1 & 0.316 & 0.323 & 0.330 & 0.366 & 0.310 \\
& 2 & 0.393 & 0.395 & 0.371 & 0.441 & 0.345 \\
& 4 & 0.508 & 0.512 & 0.501 & 0.564 & 0.467 \\
& 8 & 0.742 & 0.740 & 0.732 & 0.771 & 0.724 \\
\bottomrule
\end{tabular}
\end{table*}

\begin{table*}[t]
\centering
\small
\setlength{\tabcolsep}{5pt}
\caption{
\textbf{Top-$k$ ranking baselines.}
Values are semantic distance to the full-KG output; lower is closer to full-KG behavior. All methods improve as $k$ increases, showing that compression is robust across semantic, random, and topology-based selectors. No single selector dominates across all models and $k$ values.
}
\label{tab:ranking_baselines}
\resizebox{0.7\textwidth}{!}{
\begin{tabular}{llccccc}
\toprule
Model & $k$ & Betweenness & Degree & PageRank & Random & Semantic \\
\midrule
Llama-3.1-70B & 1 & 0.3839 & 0.4104 & 0.3836 & 0.3820 & 0.3826 \\
Llama-3.1-70B & 2 & 0.3636 & 0.3816 & 0.3825 & 0.3564 & 0.3796 \\
Llama-3.1-70B & 4 & 0.3659 & 0.3717 & 0.3667 & 0.3052 & 0.3412 \\
Llama-3.1-70B & 8 & 0.1563 & 0.1789 & 0.1786 & 0.1771 & 0.1778 \\
\midrule
Mistral-7B & 1 & 0.3642 & 0.3525 & 0.3570 & 0.3719 & 0.3973 \\
Mistral-7B & 2 & 0.3402 & 0.3413 & 0.3356 & 0.3287 & 0.3810 \\
Mistral-7B & 4 & 0.3380 & 0.3245 & 0.3334 & 0.2486 & 0.3316 \\
Mistral-7B & 8 & 0.1029 & 0.1096 & 0.1013 & 0.1075 & 0.1165 \\
\bottomrule
\end{tabular}}
\end{table*}

\subsection{Sampling Variability}
\label{app:sampling_variability}

We also test whether the observed condition effects are larger than generation noise. For a subset of problems and conditions, we generate multiple samples and compute the within-condition standard deviation of TRR. We compare this to the between-condition range in mean TRR across conditions.

Table~\ref{tab:sampling_variability} shows that between-condition differences are much larger than within-condition sampling variation. For Llama-3.1-70B, the between-condition range is $0.9100$ and the within-condition standard deviation is $0.0504$, giving a signal-to-noise ratio of $18.1\times$. For Mistral-7B, the between-condition range is $0.9167$ and the within-condition standard deviation is $0.0625$, giving a signal-to-noise ratio of $14.7\times$. This suggests that the main effects are not artifacts of single-sample generation noise. 
Tables~\ref{tab:sampling_condition_means} and~\ref{tab:sampling_condition_stds} further show per-condition mean TRR and standard deviations in the repeated-sampling check. The condition ranking remains broadly stable under repeated sampling, and within-condition variance is small relative to between-condition differences across all conditions.


\begin{table*}[!htbp]
\centering
\small
\setlength{\tabcolsep}{4pt}

\begin{minipage}[t]{0.47\textwidth}
\centering
\caption{
\textbf{Sampling variability check.}
Between-condition TRR differences are much larger than within-condition sampling noise.
}
\label{tab:sampling_variability}
\resizebox{\linewidth}{!}{
\begin{tabular}{lccc}
\toprule
Model & Between-cond. & Within-cond. & SNR \\
\midrule
Llama-3.1-70B & 0.9100 & 0.0504 & $18.1\times$ \\
Mistral-7B    & 0.9167 & 0.0625 & $14.7\times$ \\
\bottomrule
\end{tabular}}
\end{minipage}
\hfill
\begin{minipage}[t]{0.50\textwidth}
\centering
\caption{
\textbf{Mean TRR by condition in the repeated-sampling check.}
The condition ranking remains broadly stable under repeated sampling.
}
\label{tab:sampling_condition_means}
\resizebox{\linewidth}{!}{
\begin{tabular}{lccccc}
\toprule
Model & Full KG & No KG & Random KG & Targeted $k=4$ & Targeted $k=8$ \\
\midrule
Llama-3.1-70B & 0.8567 & 0.0000 & 0.8750 & 0.9100 & 0.8700 \\
Mistral-7B    & 0.8150 & 0.0000 & 0.8250 & 0.9167 & 0.8283 \\
\bottomrule
\end{tabular}}
\end{minipage}

\vspace{0.9em}

\begin{minipage}[t]{0.68\textwidth}
\centering
\caption{
\textbf{Within-condition standard deviation in the repeated-sampling check.}
Values report TRR standard deviation across repeated generations for each condition.
}
\label{tab:sampling_condition_stds}
\resizebox{\linewidth}{!}{
\begin{tabular}{lccccc}
\toprule
Model & Full KG & No KG & Random KG & Targeted $k=4$ & Targeted $k=8$ \\
\midrule
Llama-3.1-70B & 0.0609 & 0.0000 & 0.0618 & 0.0504 & 0.0788 \\
Mistral-7B    & 0.0916 & 0.0000 & 0.0761 & 0.0751 & 0.0696 \\
\bottomrule
\end{tabular}}
\end{minipage}

\end{table*}

\section{Additional Intra-Family Scaling Results}
\label{app:within_family}

The main text includes within-family checks for Llama and Mistral to test whether the compression result is only a cross-family artifact. Here we report the full Mistral-family numbers. The purpose of this analysis is not to claim a simple parameter-count scaling law. Instead, it checks whether the top-$k$ compression effect remains visible when model family is held fixed. Tables~\ref{tab:mistral_family_trr} and~\ref{tab:mistral_family_sufficiency} report TRR and semantic distance results respectively. All three Mistral models move closer to the full-KG output as $k$ increases, confirming that the compression effect is present within the family. However, the ordering is not monotonic in model size: the 22B model is not consistently closer to full-KG behavior than the 12B or 7B models.


\begin{table*}[t]
\centering
\small
\setlength{\tabcolsep}{4.5pt}
\begin{minipage}[b]{0.48\textwidth}
\centering
\caption{TRR under Mistral-family top-$k$ KG conditions. TRR is highest for small targeted graphs because the denominator contains fewer, highly salient KG objects.}
\label{tab:mistral_family_trr}
\begin{tabular}{lccc}
\toprule
Condition & Mistral-7B & Mistral-12B & Mistral-22B \\
\midrule
No KG   & 0.000 & 0.000 & 0.000 \\
Top-1   & 1.000 & 0.950 & 1.000 \\
Top-2   & 0.975 & 0.950 & 0.992 \\
Top-4   & 0.958 & 0.917 & 0.917 \\
Top-8   & 0.846 & 0.802 & 0.787 \\
Full KG & 0.717 & 0.714 & 0.675 \\
\bottomrule
\end{tabular}
\end{minipage}\hfill
\begin{minipage}[b]{0.48\textwidth}
\centering
\caption{Within-family Mistral sufficiency check. Values are semantic distance to the full-KG output; lower is closer to full-KG behavior. All three models improve as $k$ increases, but the trend is not monotonic in model size.}
\label{tab:mistral_family_sufficiency}
\begin{tabular}{lcccc}
\toprule
Model & $k=1$ & $k=2$ & $k=4$ & $k=8$ \\
\midrule
Mistral-7B  & 0.3705 & 0.3700 & 0.3360 & 0.2499 \\
Mistral-12B & 0.3772 & 0.3725 & 0.3484 & 0.2456 \\
Mistral-22B & 0.3690 & 0.3765 & 0.3550 & 0.2567 \\
\bottomrule
\end{tabular}
\end{minipage}
\end{table*}

\section{Topology and Semantic-Role Ablations}
\label{app:topology_semantics}

To test whether important triples are identified by graph topology alone, we compare removals of bridge, peripheral, and random triples. Bridge triples provide a simple topology-driven baseline: they connect major parts of the problem graph and should be important if graph position alone determines influence. The results do not support a bridge-only explanation. Peripheral and random removals often produce comparable or larger semantic shifts than bridge removals, especially for Gemini and Llama-3.1-70B. This suggests that triple importance is not reducible to simple graph position. Table~\ref{tab:knockout_sem_dist} summarizes the knockout semantic-distance results. The relation-type analysis gives a clearer signal: outcome-facing and task-relevant triples often produce larger changes than bridge status alone.

\begin{table*}[!htbp]
\centering
\small

\begin{minipage}[t]{0.50\textwidth}
\centering
\caption{
\textbf{Semantic distance after triple knockout by removal type.}
Peripheral and random removals are often as disruptive as bridge removals, showing that graph position alone does not explain triple importance.
}
\label{tab:knockout_sem_dist}
\begin{tabular}{lccc}
\toprule
Model & Bridge & Peripheral & Random \\
\midrule
Gemini 2.5 Flash        & 0.0747 & 0.1027 & 0.0941 \\
Llama-3.1-70B & 0.0306 & 0.0579 & 0.0366 \\
Mistral-7B    & 0.0425 & 0.0537 & 0.0493 \\
\bottomrule
\end{tabular}
\end{minipage}
\hfill
\begin{minipage}[t]{0.43\textwidth}
\centering
\caption{
\textbf{Semantic-distance sensitivity to embedding encoder.}
Both encoders preserve the same sufficiency trend: top-$k$ outputs move closer to the full-KG output as $k$ increases. The two distance series are strongly correlated (Spearman $\rho=0.965$).
}
\label{tab:encoder_sensitivity}
\begin{tabular}{lcc}
\toprule
$k$ & MPNet-base & MiniLM-L6 \\
\midrule
1 & 0.3053 & 0.3650 \\
2 & 0.3014 & 0.3587 \\
4 & 0.2607 & 0.3171 \\
8 & 0.0379 & 0.0416 \\
\bottomrule
\end{tabular}
\end{minipage}

\end{table*}


\subsection{Semantic Role Versus Topological Centrality}
\label{app:topological_baselines}

To test whether graph importance is explained by topology alone, we compare the effect of removing three types of triples: bridge triples, peripheral triples, and randomly selected triples. Bridge triples are selected based on graph position, while peripheral triples are selected from less central graph locations. Random removals provide a baseline for removal effects not tied to either semantic role or graph centrality. Table~\ref{tab:topology_knockout} summarizes the full knockout results with per-model TRR and $\Delta$TRR alongside semantic distance. The results do not support a simple topological-centrality explanation. Removing bridge triples changes outputs, and for Gemini 2.5 Flash it slightly reduces TRR. However, peripheral and random removals can be as disruptive or more disruptive in semantic distance. This indicates that graph position alone does not identify the triples most important for generation. The relation-type analysis in the main paper gives a clearer signal: outcome-facing and task-relevant triples often produce larger changes than bridge status alone. This result motivates our main claim that the compressed useful subset is semantically organized. Topological centrality is informative, but it is not sufficient. In future work, one could compare the relevance ranking used here against explicit degree, betweenness, and PageRank rankings. Betweenness centrality measures how often a node lies on shortest paths between other nodes, while degree centrality and PageRank capture different notions of graph importance. These baselines are useful, but our current knockout results suggest that scientific role is a stronger predictor of output disruption than simple graph position.

\begin{table*}[t]
\centering
\small
\setlength{\tabcolsep}{5pt}
\caption{Semantic distance shows that peripheral and random removals can be as disruptive as bridge removals, indicating that graph topology alone does not explain triple importance. Gemini 2.5 Flash refers to Gemini 2.5 Flash}
\label{tab:topology_knockout}
\begin{tabular}{llrrrr}
\toprule
Model & Removal type & Count & Mean semantic distance & TRR & $\Delta$TRR \\
\midrule
Gemini 2.5 Flash & None       & 300 & 0.0000 & --     & -- \\
Gemini 2.5 Flash & Bridge     & 300 & 0.0747 & 0.6081 & -0.0216 \\
Gemini 2.5 Flash & Peripheral & 300 & 0.1027 & 0.6652 & +0.0356 \\
Gemini 2.5 Flash & Random     & 297 & 0.0941 & 0.6489 & +0.0192 \\
\midrule
Llama-70B & None       & 300 & 0.0000 & --     & -- \\
Llama-70B & Bridge     & 300 & 0.0306 & 0.6538 & -0.0007 \\
Llama-70B & Peripheral & 300 & 0.0579 & 0.6912 & +0.0367 \\
Llama-70B & Random     & 297 & 0.0366 & 0.6809 & +0.0264 \\
\midrule
Mistral-7B & None       & 300 & 0.0000 & --     & -- \\
Mistral-7B & Bridge     & 300 & 0.0425 & 0.7340 & +0.0469 \\
Mistral-7B & Peripheral & 300 & 0.0537 & 0.7495 & +0.0624 \\
Mistral-7B & Random     & 297 & 0.0493 & 0.7645 & +0.0774 \\
\bottomrule
\end{tabular}
\end{table*}

\section{KG Size and Context Length}
\label{app:kg_size}

The targeted condition can have a larger length proxy than some full-path variants because selected triples are verbalized with more explicit natural-language context. We therefore treat this as a context-length proxy. Table~\ref{tab:kg_size_stats} summarizes per-problem KG size statistics. Each problem has a compact local KG concentrated around 16 triples, meaning top-8 uses approximately half of the full local graph and top-4 uses about one quarter. Table~\ref{tab:triple_count_distribution} shows the distribution of triple counts across problems, confirming that local KGs are tightly concentrated. Table~\ref{tab:context_length_proxy} reports the mean verbalized context length for each KG condition; these values are shown as prompt-length proxies rather than triple counts.

\section{Encoder Sensitivity for Semantic Distance}
\label{app:encoder_sensitivity}

Throughout this study, we used MiniLM-6 sentence transformer. Here, we recompute semantic distance using both MPNet-base and MiniLM-L6 sentence encoders without rerunning generation. Absolute distances differ across encoders, but the qualitative sufficiency conclusion is unchanged: top-$k$ outputs move closer to full-KG outputs as $k$ increases. The distance curves are show Spearman $\rho=0.965$, suggesting that the compression result is not an artifact of a single embedding model. This shows that semantic distance is used only as a behavioral similarity measure, not as a metric to evaluate hypothesis quality. If the compression trend depended on one encoder, the sufficiency result would be less reliable. Table~\ref{tab:encoder_sensitivity} reports these values.


\begin{table*}[t]
\centering
\small
\caption{
\textbf{Per-problem KG size.}
Each problem has a compact local KG, with most problems containing 16 triples. Top-8 therefore uses about half of the full local graph, while top-4 uses about one quarter.
}
\label{tab:kg_size_stats}
\begin{tabular}{lc}
\toprule
Statistic & Value \\
\midrule
Shared KG nodes & 6,641 \\
Shared KG edges & 8,058 \\
Mean triples/problem & 16.1 \\
Median triples/problem & 16.0 \\
Std. dev. & 0.5 \\
Min--max & 15--18 \\
25th--75th percentile & 16--16 \\
Top-8 fraction of full KG & 49.7\% mean; 50.0\% median \\
Top-4 fraction of full KG & 24.8\% mean \\
\bottomrule
\end{tabular}
\end{table*}



\begin{table*}[!htbp]
\centering
\small

\begin{minipage}[t]{0.42\textwidth}
\centering
\caption{
\textbf{Distribution of per-problem triple counts.}
The local KGs are tightly concentrated around 16 triples per problem.
}
\label{tab:triple_count_distribution}
\begin{tabular}{lr}
\toprule
Triple-count bin & Number of problems \\
\midrule
1--4 & 0 \\
5--8 & 0 \\
9--12 & 0 \\
13--16 & 415 \\
17--20 & 85 \\
21--30 & 0 \\
31+ & 0 \\
\bottomrule
\end{tabular}
\end{minipage}
\hfill
\begin{minipage}[t]{0.48\textwidth}
\centering
\caption{
\textbf{Context length proxy by KG condition.}
These values summarize the approximate verbalized context length in the main pipeline. They should be interpreted as prompt-length proxies rather than triple counts.
}
\label{tab:context_length_proxy}
\begin{tabular}{lr}
\toprule
Condition & Mean length proxy \\
\midrule
Baseline & 0.0 \\
Sparse & 575.0 \\
Medium & 1065.0 \\
2-hop & 1065.0 \\
T1 coarse & 1067.0 \\
T3 multihop & 1069.0 \\
Full path & 1070.0 \\
Shuffled & 1085.0 \\
Random & 1086.0 \\
Targeted & 1524.0 \\
Dense & 2589.0 \\
\bottomrule
\end{tabular}
\end{minipage}

\end{table*}

\end{document}